\documentclass[11pt,letterpaper]{article}
\usepackage{styles/preprint}

\hypersetup{
  pdftitle={The Data Debt of Stockouts: Partial Identification and Minimax-Regret Inventory Control},
  pdfauthor={Jinbiao Chen; Shuang Jin; Hanzhang Qin},
  pdfkeywords={inventory learning, censored demand, partial identification}
}

\title{The Data Debt of Stockouts: Partial Identification and Minimax-Regret Inventory Control}
\author{%
  {\large Jinbiao Chen\textsuperscript{1,\textdagger}\quad
   Shuang Jin\textsuperscript{1,\textdagger}\quad
   Hanzhang Qin\textsuperscript{1,2,*}}\par
  \vspace{9pt}
  {\small
  \textsuperscript{1}Department of Industrial Systems Engineering and Management,\par
  National University of Singapore\par
  \vspace{4pt}
  \textsuperscript{2}Institute of Operations Research and Analytics,\par
  National University of Singapore\par
  \vspace{7pt}
  \href{mailto:bill.cjb@nus.edu.sg}{\texttt{bill.cjb@nus.edu.sg}}\quad
  \href{mailto:jinshuang.7@foxmail.com}{\texttt{jinshuang.7@foxmail.com}}\quad
  \href{mailto:hzqin@nus.edu.sg}{\texttt{hzqin@nus.edu.sg}}\par
  \vspace{6pt}
  \textsuperscript{\textdagger}Equal contribution\quad
  \textsuperscript{*}Corresponding author\par}}
\date{}

\begin{document}
\maketitle
\begin{abstract}
\noindent\textbf{Problem definition:} Inventory decisions select which demand enters sales logs. Unrecorded forecasts can leave demand frequencies unidentified even when stocking levels are recorded. We study which multistage ordering decisions these logs can justify and what finite samples can guarantee.

\noindent\textbf{Methodology/results:} Nested stockout thresholds yield exact sets of compatible demand laws and finite representations for policy evaluation. We choose one deterministic inventory rule to minimize its largest expected cost gap from the informed optimum under the same law. This population value is data debt for the specified information and policy class. Under independent demand, carryover can create positive debt despite common optimal actions in every isolated period. Finite samples add learning error above debt. The required sample count grows linearly with the selection bound, with upper and lower orders matching up to logarithmic factors. This rate requires fixed demand and cost bounds, sufficient revelation, and a uniformly positive probability of clearing all feasible stock across the confidence sets. Controlled and semi-synthetic studies separate policy improvements from tighter guarantees.

\noindent\textbf{Managerial implications:} More records can tighten guarantees while debt persists. Deeper revelation can establish common optimality without fully identifying demand. Using these guarantees requires validated logging, a credible selection range, and a regret tolerance in cost units. Information acquisition must distinguish tighter assurance from operating savings.
\end{abstract}


\noindent\textbf{Subject classifications:} Inventory/production: learning from stockout logs. Statistics: informative censoring. Decision analysis: minimax regret.

\section{Introduction}\label{sec:intro}

Demand censoring is common in supply chains and logistics. Firms observe sales or deliveries but not all demand because stock and capacity limit what can be served \citep{godfrey2001}. Sales below the available stock reveal demand exactly. A sellout reveals that demand reached the recorded boundary but not how far it exceeded it.

Our log records sales and stock available before demand. A manager may stock more when an unrecorded forecast predicts a busy day, making high-stock days unrepresentative. The sensitivity bound $\Gamma$ limits the difference between revealed and hidden demand compositions. Each period's demand distribution stays stable between learning and deployment.

\begin{figure}[t]
\FIGURE
{\includegraphics[width=\linewidth]{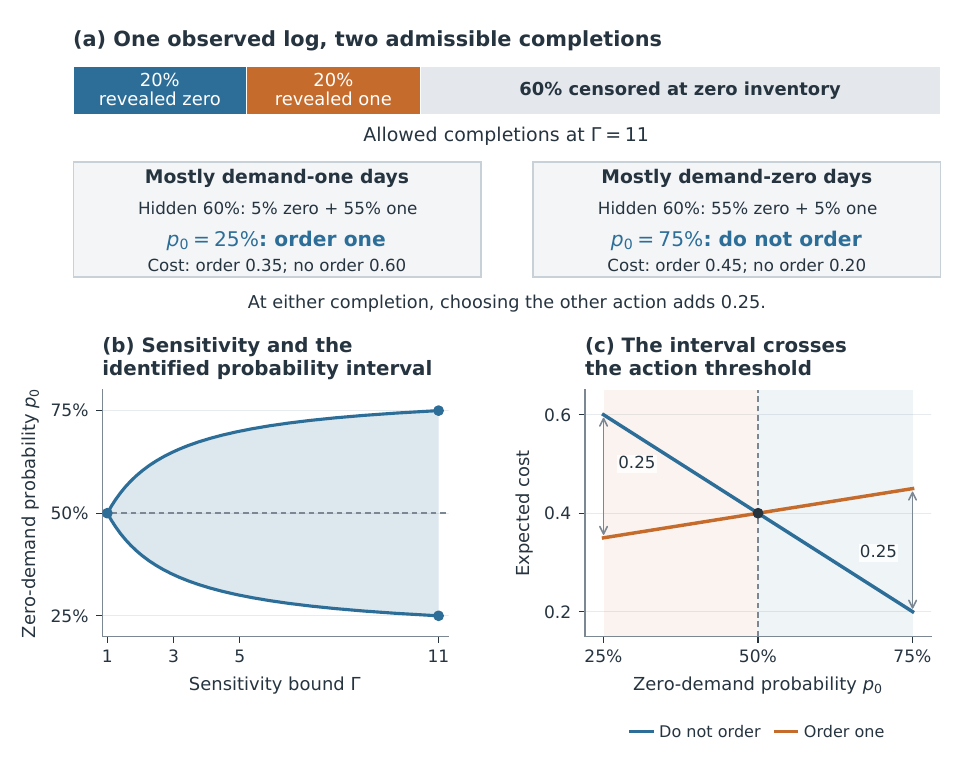}}
{The same stockout log supports different demand frequencies and opposite optimal actions. (a) Two compatible completions at $\Gamma=11$. Percentages refer to the full population. (b) The selection bound varies while the log stays fixed. (c) Ordering and no-order costs. The actions tie at $p_0=0.5$, and the other action at either endpoint adds 0.25 in expected cost.\label{fig:intro-same-log}}
{}
\end{figure}

To see how unrecorded demand can change an inventory decision, consider an item with demand zero or one. Its long-run log contains 20\% revealed zeros, 20\% revealed ones, and 60\% censored at zero inventory. The revealed zero-to-one odds are 1:1. A selection bound $\Gamma=11$ allows hidden odds from 1:11 to 11:1, giving overall zero-demand probabilities $p_0$ from 0.25 to 0.75 in Figure~\ref{fig:intro-same-log}(a). With setup, holding, and shortage costs of 0.30, 0.20, and 0.80 and no unit ordering cost, the endpoint explanations recommend opposite actions. Figure~\ref{fig:intro-same-log}(c) shows that the wrong action costs an extra 0.25 in expectation. Both values appear in uncensored records, but their population frequencies remain unresolved.

A deterministic rule specifies one order for each period and inventory level. Its regret is the expected cost above the best such rule knowing the demand law. We minimize the largest regret across compatible laws. The resulting population data debt is 0.25 in the opening example, even with exact log frequencies. Carryover links decisions through leftover stock: isolated periods can share optimal actions while the full horizon has no common optimal policy.

Our analysis proceeds from demand identification to common ordering rules and then to finite-sample guarantees. Here $H$ is the horizon, $B$ bounds integer demand and inventory, and $N$ counts independent and identically distributed (iid) historical trajectories, each recording an $H$-period cycle.
\begin{enumerate}[label=\arabic*.,leftmargin=1.6em,itemsep=0.3em,parsep=0pt,topsep=0.4em]
\item \emph{Exact demand identification from nested thresholds.} Theorem~\ref{thm:sharp} characterizes the demand laws that reproduce the log under hidden selection. Higher stocking levels reveal demand beyond lower ones, giving the records a nested structure. Proposition~\ref{prop:direct-demand} uses this structure to eliminate unobserved allocations of censored demand, leaving $B+1$ demand probabilities and $O(B)$ linear constraints. Proposition~\ref{prop:endpoint-hazard} gives at most $2^B$ distributions from lower or upper stopping probabilities; their weighted averages fill the demand set. Corollary~\ref{cor:recorded-boundary} establishes equivalence with the full periodwise sales-and-stocking record. Additional historical stock-flow restrictions can narrow this set.

\item \emph{A carryover mechanism for multistage data debt.} Using each period's log distribution, Proposition~\ref{prop:reduced} identifies the smallest worst-case regret attainable by one deterministic rule. Proposition~\ref{prop:debt} shows that this debt vanishes exactly when compatible models share an optimal rule. Proposition~\ref{prop:homogeneous} then isolates the carryover mechanism: with identical costs across periods, both isolated periods can have zero debt while the full horizon has positive debt.

\item \emph{Explicit sensitivity dependence in learning above debt.} Finite logs add uncertainty about the demand set. Enlarging it for sampling error gives an upper regret bound, or certificate (Theorems~\ref{thm:pac}--\ref{thm:contraction}). We implement this step with Clopper--Pearson (CP) binomial confidence intervals. A second construction uses demand hazards: the probabilities of stopping at a level conditional on reaching it. Theorems~\ref{thm:gamma-upper}--\ref{thm:gamma-lower} and Corollary~\ref{cor:gamma-sample} give the matching trajectory order $N=\widetilde\Theta(\Gamma H^2/(q\varepsilon^2))$, up to logarithms, in the stated fixed-confidence and small-error regime. Here $\varepsilon$ bounds excess worst-case regret above debt, and $q$ bounds the chance a record reaches a threshold and reveals whether demand stops there. The upper bound holds with fixed demand and cost bounds, uniformly positive stock-clearing probability across confidence sets, and exact hazard-set optimization. The lower family has positive debt.
\end{enumerate}

The numerical studies compare hazard and CP, trace cost conditions that change orders, and examine common optimality under deeper revelation. Section~\ref{sec:historical-observation} describes a compatible logging architecture, and Section~\ref{sec:discussion} explains how a cost-unit tolerance can guide deployment and information collection.

\section{Related Literature}\label{sec:literature}

\paragraph{Learning with observed and censored demand.}
\citet{levi2015} bound empirical newsvendor performance, and \citet{besbes2023data} characterize worst-case performance across sample sizes. Related learning guarantees include \citet{xie2026vc,zhang2024,fan2026,lyu2025metapolicy}. Censored-demand learning uses threshold hazards, stochastic approximation, and adaptive ordering \citep{godfrey2001,huh2009,huh2011,besbes2013,ban2020,yuan2021,lyu2024,ding2024}. In the single-period censored newsvendor, \citet{hssaine2024} derive irreducible regret and statistical guarantees for distributions agreeing below the largest historical stock. \citet{kumar2026} study worst-case performance and targeted exploration.

In offline pricing, \citet{bu2023offline} characterize when censored sales permit near-optimal decisions and attain the best-achievable minimax loss otherwise. We characterize the exact demand set under bounded informative stocking and its consequences for multistage ordering with carryover.

Our closest multistage comparisons, \citet{qin2026information} and \citet{han2026biased}, also observe sales and stocking boundaries. \citet[Sections~2.2 and~5]{qin2026information} fix period demand laws conditional on the complete boundary array; coverage for ordering and carryover enables policy learning even when absolute costs remain unidentified. \citet{han2026biased} allow history-dependent stocking with current demand laws fixed conditional on past observations. Their biased sample-average approximation (SAA) learns from offline coverage and generates it online. Here stocking may use omitted current-demand predictors, leaving frequencies within observed support unresolved. Planning demand remains independent; policies are precommitted and deterministic. Section~\ref{sec:gamma-sample} distinguishes our explicit $\Gamma$ dependence above positive debt from the cubic scalar-record horizon order in \citet{qin2026information}.

\paragraph{Sensitivity to hidden selection.}
\citet{kaciroti2012} compare event odds across observed and missing-data patterns in a Bayesian product-limit model. Here threshold odds bounds between revealed and hidden groups, combined with nested stockout records, yield linear demand constraints and an endpoint recursion.

Odds-ratio sensitivity, confounding-robust policy learning, and sequential evaluation provide established tools \citep{tan2006,zhao2019,yadlowsky2022,kallus2021,namkoong2020}. \citet{hess2026} derive sharp value bounds, efficient estimators, and learning guarantees, including baseline improvement. \citet[Appendix~B.2]{jin2026policy} minimize sensitivity-set regret against the same-law class optimum, allowing deployment randomization. We build on this criterion. The inventory contribution links nested censored records to an exact marginal set, a common-law carryover problem, and explicit sensitivity-dependent finite-sample limits. Table~\ref{ec-tab:technical-comparison} separates the information restrictions and targets.

\paragraph{Inventory decisions under partial information.}
Minimax regret is established in inventory \citep{perakis2008} and statistical decisions under partial identification \citep{stoye2007,stoye2012,montielolea2026}. Unlike moment- or distance-based uncertainty sets \citep{delage2010,esfahani2018}, our demand laws must have a stocking mechanism that reproduces the retained record. A sharp demand set retains every law compatible with the historical model and excludes every incompatible law; additional joint observations or stock-flow constraints change that benchmark. Dynamic and learning inventory models examine procurement commitments, alternative feedback, and demand dependence \citep{wang2017commitments,bekci2023,chen2024,katehakis2024,mukherjee2026,gundem2025,tang2025}; revenue-management unconstraining supplies demand completions \citep{talluri2004}.

\section{Model Setup}\label{sec:model}

We first define future inventory decisions and their cost, then describe the historical logs used to learn a rule. Historical stocking may use demand information absent from those logs.

\subsection{The planning problem}

There are $H$ periods, and inventory is counted in integer units up to capacity $B$. Before period $t$'s order, the stock is $X_t\in\{0,\ldots,B\}$. Initial stock $X_1=x_1$ is fixed, with $x_1=0$ in the numerical studies. The decision $Z_t\in\{X_t,\ldots,B\}$ is the stock level \emph{after} ordering. The order quantity is $Z_t-X_t$. Demand $D_t\in\{0,\ldots,B\}$ then arrives, sales are the smaller of demand and available stock, and leftover stock carries to the next period. Unmet demand is lost rather than backlogged.

\begin{equation}
X_{t+1}=Z_t-\min\{D_t,Z_t\},\qquad
 g_t(x,z,d)=K_t\ind\{z>x\}+c_t(z-x)+h_t(z-d)^++b_t(d-z)^+.
\label{eq:dynamics-cost}
\end{equation}

The four nonnegative coefficients charge for placing an order ($K_t$), each unit ordered ($c_t$), leftover stock ($h_t$), and unmet demand ($b_t$). $a^+=\max\{a,0\}$ and $\ind\{E\}$ is one when event $E$ holds and zero otherwise. A policy $\pi=(\pi_1,\ldots,\pi_H)$ is a lookup table with $\pi_t(x)\in\{x,\ldots,B\}$. This deterministic Markov policy uses only the period and current inventory. These policies form the finite class $\PiM$. Write $P_t$ for the demand distribution in period $t$ and $p_t=(p_{t,0},\ldots,p_{t,B})$ for its probability vector, where $p_{t,k}=P_t(D_t=k)$. Independence gives the joint law $P=\bigotimes_tP_t$, called a product law. $\E_P$ denotes expectation under that law.

\begin{equation}
J_P(\pi)=\E_P\!\left[\sum_{t=1}^Hg_t(X_t,\pi_t(X_t),D_t)\right],\qquad
J_P^*=\min_{\pi\in\PiM}J_P(\pi),\quad \Reg_P(\pi)=J_P(\pi)-J_P^*.
\label{eq:regret}
\end{equation}

Thus $J_P(\pi)$ is the expected cost of a proposed rule, $J_P^*$ is the informed optimum, or oracle value, in the same class, and $\Reg_P(\pi)$ is their nonnegative difference in cost units per modeled horizon. For a fixed action cap $B$, grouping larger physical demands at $B$ (\emph{top coding}) preserves inventory paths and regret after a common cost subtraction. Section~\ref{ec-sec:prelim} proves this reduction. For the perturbation proofs, we also subtract a policy-invariant component from the stage cost:

\begin{equation}
\widetilde g_t(x,z,d)=K_t\ind\{z>x\}+c_t(z-x)+h_tz-(h_t+b_t)\min\{d,z\}.
\label{eq:transformed-cost}
\end{equation}

Let $\widetilde J_P(\pi)=\E_P[\sum_{t=1}^H\widetilde g_t(X_t,\pi_t(X_t),D_t)]$ and $\widetilde J_P^*=\min_{\pi\in\PiM}\widetilde J_P(\pi)$. Because $g_t=\widetilde g_t+b_td$, the transformed and original regrets coincide. The subtraction removes a demand-dependent cost that no policy can affect, so later probability perturbations need only track the policy-sensitive part of cost.

\subsection{The historical observation}\label{sec:historical-observation}

The baseline historical file records both sales $Y_t$ and the effective censoring level $C_t\in\{0,\ldots,B+1\}$. The historical observation model imposes no interperiod stock-flow restriction on $C_{1:H}=(C_1,\ldots,C_H)$. We call $C_t$ the \emph{reveal depth}. Demand beyond that level is not distinguished by the record, while depth $B+1$ reveals the full modeled range. The revelation indicator $\Delta_t$ is determined by $(Y_t,C_t)$. It equals one when $Y_t<C_t$ and zero at a sell-through tie. The probabilities $o_{t,d}$ and $s_{t,c}$ summarize revealed and censored records, respectively:

\begin{equation}
Y_t=\min\{D_t,C_t\},\qquad \Delta_t=\ind\{D_t<C_t\},\qquad
 o_{t,d}=\Prb(Y_t=d,\Delta_t=1),\quad s_{t,c}=\Prb(Y_t=c,\Delta_t=0).
\label{eq:observation}
\end{equation}

Write $O_t^C=(Y_t,\Delta_t,C_t)$ for the recorded triple and $Q_t^C$ for its population law. The coarsened record $O_t=(Y_t,\Delta_t)$ has law $Q_t$, summarized by $o_t=(o_{t,0},\ldots,o_{t,B})$ and $s_t=(s_{t,0},\ldots,s_{t,B})$. Sales of four reveal demand of four when stock remains, but imply only $D_t\ge4$ at a sellout. We use $(o_t,s_t)$ to construct demand sets. Corollary~\ref{cor:recorded-boundary} shows why these summaries suffice for periodwise identification under the selection restriction below. Files retaining only sales and a validated revelation indicator use the same construction. A physical capacity or stock proxy must be validated as the effective boundary before it can be treated as $C_t$ (Section~\ref{ec-sec:depth-information}).

A pre-demand signal can predict current demand despite period independence (Section~\ref{ec-sec:timing}). One illustrative use learns a fixed rule from historical cycles allowing leftover stock to be returned, reallocated, or removed, then deploys with carryover. Demand marginals stay stable across these protocols. Without removals, $C_{t+1}\ge C_t-Y_t$ adds constraints; combining periodwise demand sets then gives a conservative benchmark, as Section~\ref{sec:rectangular} explains.

At each threshold $d<B$, consider records with $D_t\ge d$. Those with $C_t>d$ reveal whether demand stops at $d$ or exceeds it. Those with $C_t\le d$ do not. In either group, the stopping probability is its demand \emph{hazard}, and its \emph{odds} are that probability divided by the probability of exceeding $d$. The \emph{hidden-selection bound} $\Gamma\ge1$ limits the unrevealed-to-revealed odds ratio:
\[
\frac{1}{\Gamma}\le
\frac{\Prb(D_t=d,C_t\le d)/\Prb(D_t>d,C_t\le d)}
     {\Prb(D_t=d,C_t>d)/\Prb(D_t>d,C_t>d)}
\le\Gamma.
\]
Cross-multiplication covers zero probabilities and leaves empty-group comparisons unrestricted. Thus $\Gamma=1$ equates the hazards when both groups have positive mass. Full demand--stock independence is sufficient but not necessary. Figure~\ref{fig:intro-same-log}(b) shows how relaxing this bound widens the compatible demand interval.

The censoring proportion $\Prb(\Delta_t=0)=\sum_cs_{t,c}$ measures how often demand is hidden. The bound $\Gamma$ limits how different its composition may be. Independent stocking can cause frequent censoring at $\Gamma=1$. Varying $\Gamma$ at a fixed log changes the permitted selection, while its censoring frequencies and depths remain fixed.

\subsection{Population and deployment assumptions}

The assumptions below collect the observation, demand, sensitivity, and deployment requirements. Independent historical trajectories are additionally required for the finite-sample guarantees in Section~\ref{sec:finite-sample}.

\begin{assumption}[Observation architecture]\label{ass:A1}
The baseline historical record contains $(Y_t,C_t)$ and follows \eqref{eq:observation}, with sell-through ties treated as censored. Its derived indicator is $\Delta_t=\ind\{Y_t<C_t\}$.
\end{assumption}
Identification and certification use a validated effective boundary and revelation indicator.

\begin{assumption}[Planning demand]\label{ass:A2}
Demand has support $\{0,\ldots,B\}$ and is independent across periods, with potentially period-specific marginals. There is no terminal cost.
\end{assumption}
Independence makes periodwise marginals sufficient for multistage cost evaluation. It does not require $D_t\perp C_t$. With serial dependence, the same marginals can give different policy costs. 
\begin{assumption}[Sensitivity]\label{ass:A3}
The historical law satisfies the odds bound defined in Section~\ref{sec:historical-observation} at the stated $\Gamma\ge1$ on the full prespecified universe $\mathcal U=\{1,\ldots,H\}\times\{0,\ldots,B-1\}$.
\end{assumption}
Calibrating $\Gamma$ requires substantive knowledge, external information, or validated deeper revelation (Section~\ref{ec-sec:calibration}). Understating it can exclude the true law.

\begin{assumption}[Deployment class]\label{ass:A4}
The learner commits before deployment to one policy in $\PiM$.
\end{assumption}
The informed comparator uses the same class. A deployment mixture draws one of several rules before use. Such mixtures and rules using additional sales history define different benchmarks.

\begin{assumption}[Population stability]\label{ass:A5}
For each period, the deployment demand marginal equals the latent demand marginal associated with historical observation law $Q_t$.
\end{assumption}
A5 allows stocking changes while holding demand marginals fixed. Population shifts require new data or a shift model. Section~\ref{sec:semisynthetic} checks alternative post-cutoff evaluation laws.

\section{From Demand Identification to Inventory Decisions}\label{sec:identification}

We first treat log probabilities as known and identify the demand laws that explain them. We then simplify this set for policy evaluation, combine periods, and ask whether one rule is optimal throughout. Finite-sample uncertainty enters in Section~\ref{sec:guarantees}.

\subsection{Demand laws supported by the log}\label{sec:threshold-model}

A record censored at stock $c$ tells us only that demand was at least $c$. We distribute that record's probability among possible demands $k\ge c$, writing $m_{t,k,c}$ for the amount assigned to each. These assignments must reproduce the recorded censored frequencies. Adding them to revealed demand gives a candidate demand distribution:

\begin{equation}
\begin{gathered}
m_{t,k,c}=\Prb(D_t=k,C_t=c),\qquad 0\le c\le k\le B,\\
m_{t,k,c}\ge0,\qquad \sum_{k=c}^{B}m_{t,k,c}=s_{t,c},\qquad
p_{t,k}=o_{t,k}+\sum_{c=0}^km_{t,k,c}.
\end{gathered}
\label{eq:latent}
\end{equation}

We now express the hidden-selection bound from Section~\ref{sec:historical-observation} through these allocations. The revealed stopping and continuing masses are $e_{t,d}=o_{t,d}$ and $\tau_{t,d}=\sum_{y>d}(o_{t,y}+s_{t,y})$. For records hidden at threshold $d$, $u_{t,d}$ is the probability mass stopping at $d$, and $v_{t,d}$ is the mass continuing beyond it:

\begin{equation}
u_{t,d}=\sum_{c=0}^dm_{t,d,c},\qquad
v_{t,d}=\sum_{k=d+1}^B\sum_{c=0}^dm_{t,k,c}.
\label{eq:hidden-masses}
\end{equation}

The revealed and unrevealed odds are $e_{t,d}/\tau_{t,d}$ and $u_{t,d}/v_{t,d}$. Cross-multiplying the bound gives
\begin{align}
e_{t,d}v_{t,d}&\le\Gamma\tau_{t,d}u_{t,d},\label{eq:gamma1}\\
\tau_{t,d}u_{t,d}&\le\Gamma e_{t,d}v_{t,d}.\label{eq:gamma2}
\end{align}

The observed masses $e_{t,d},\tau_{t,d}$ are constants, while $u_{t,d},v_{t,d}$ are linear in the allocations, so the sensitivity restrictions are linear.

Let $\Mset_{\Gamma,t}(o_t,s_t)$ collect all allocations satisfying \eqref{eq:latent}--\eqref{eq:gamma2}. Keeping only the resulting demand probabilities gives $\Pset_{\Gamma,t}(o_t,s_t)$, the demand-marginal \emph{projection}. A projection here means discarding the allocation variables after checking that at least one valid allocation exists. Because the restrictions are linear, the demand set is a polytope, a bounded region described by linear constraints.

\begin{theorem}[Periodwise-sharp identification]\label{thm:sharp}
$\Pset_{\Gamma,t}(o_t,s_t)$ is exactly the set of demand marginals generated by joint laws of $(D_t,C_t)$ that reproduce $(o_t,s_t)$ and satisfy the threshold-specific $\Gamma$ restrictions.
\end{theorem}

Each feasible allocation supplies a compatible historical law, and conversely. Having identified the set, we next ask whether recording each censoring level narrows it.

\begin{corollary}[Recorded-boundary equivalence]\label{cor:recorded-boundary}
Let $Q_t^C$ be a law of $(Y_t,\Delta_t,C_t)$ consistent with the observation architecture in \eqref{eq:observation}, and let $(o_t,s_t)$ be its coarsened probabilities. Under the threshold-specific $\Gamma$ restrictions, with no additional restrictions on the joint law of $(D_t,C_t)$, the demand set reproducing the full periodwise record satisfies
\[
\Pset_{\Gamma,t}^{C}(Q_t^C)=\Pset_{\Gamma,t}(o_t,s_t).
\]
\end{corollary}

The full record fixes each revealed cell $a_{t,k,c}=\Prb(Y_t=k,\Delta_t=1,C_t=c)$ for $k<c$, with row sum $o_{t,k}$. Appending a feasible hidden allocation $m$ reproduces $Q_t^C$ while preserving the demand marginal and threshold odds restrictions. Section~\ref{ec-sec:latent} proves the result, including zero-mass cases. This is a population-identification result; finite-sample confidence sets are constructed in Section~\ref{sec:finite-sample}. Propositions~\ref{prop:direct-demand} and~\ref{prop:endpoint-hazard} therefore represent the same set for both the full periodwise record and its summary.

The case $\Gamma=1$ connects the identified set to product-limit estimation, which combines successive threshold hazards into a demand distribution.

\begin{proposition}[Product-limit benchmark]\label{prop:gamma1}
If $r_{t,d}=e_{t,d}+\tau_{t,d}>0$ for every $d<B$, then at $\Gamma=1$ the demand hazard equals $e_{t,d}/r_{t,d}$, and $\Pset_{1,t}$ is the singleton discrete product-limit law.
\end{proposition}

The hazards determine the discrete product-limit, or Kaplan--Meier, law used in censored inventory learning \citep{huh2011,lyu2024}. A threshold with no revealed risk cannot be learned merely by setting $\Gamma=1$. The proof is in Section~\ref{ec-sec:latent}.

\subsection{From latent allocations to endpoint distributions}

The latent allocation proves which demand laws are possible, but its unobserved cells need not all appear in the optimization. Because a censored depth can contribute only to larger demands, cumulative tail constraints suffice to decide whether a proposed marginal admits an allocation. This yields a direct representation in demand probabilities. Suppress time subscripts and write $\tau_d=\sum_{j>d}(o_j+s_j)$.

\begin{proposition}[Direct demand-marginal representation]\label{prop:direct-demand}
$\Pset_\Gamma(o,s)$ consists exactly of probability vectors satisfying

\begin{align}
p_k&\ge o_k, &&k=0,\ldots,B,\label{eq:direct-lower}\\
\sum_{k=d}^B(p_k-o_k)&\ge\sum_{c=d}^Bs_c,&&d=1,\ldots,B,\label{eq:direct-transport}\\
o_d\left(\sum_{k>d}p_k-\tau_d\right)&\le\Gamma\tau_d(p_d-o_d),&&d<B,\label{eq:direct-gamma-lower}\\
\tau_d(p_d-o_d)&\le\Gamma o_d\left(\sum_{k>d}p_k-\tau_d\right),&&d<B,\label{eq:direct-gamma-upper}
\end{align}

together with $\sum_kp_k=1$. The projected set has $B+1$ variables and $O(B)$ constraints.
\end{proposition}

The first two constraint groups preserve revealed demand and assign censored mass only at or above its reveal depth. The last two impose the selection bound. The hazard construction below establishes sufficiency and the endpoint representation (Section~\ref{ec-sec:direct-projection}).

The same nesting gives a constructive description. At threshold $d$, let $a_d=u_d+v_d=\Prb(D\ge d,C\le d)$ be the unresolved probability mass whose revelation has ended by $d$ and whose demand reaches $d$. Its fraction stopping at $d$ is the unrevealed hazard $\nu_d=\Prb(D=d\mid D\ge d,C\le d)$. The remaining fraction continues to the next threshold, where mass newly censored there is added. When $o_d,\tau_d>0$, this gives

\begin{equation}
\begin{gathered}
[\underline\nu_d,\overline\nu_d]=\left[\frac{o_d}{o_d+\Gamma\tau_d},\frac{\Gamma o_d}{\tau_d+\Gamma o_d}\right],\qquad a_0=s_0,\\
p_d=o_d+\nu_da_d,\qquad a_{d+1}=(1-\nu_d)a_d+s_{d+1},\qquad p_B=o_B+a_B.
\end{gathered}
\label{eq:hazard-recursion}
\end{equation}

The interval is $[0,1]$ when both revealed masses are zero, $[0,0]$ when only $o_d=0$, and $[1,1]$ when only $\tau_d=0$.

The convex hull of endpoint distributions consists of all their weighted averages. The next result shows that these averages recover the entire demand set.

\begin{proposition}[Endpoint-hazard representation]\label{prop:endpoint-hazard}
$\Pset_\Gamma(o,s)$ is the convex hull of distributions generated by choosing each $\nu_d$ at $\underline\nu_d$ or $\overline\nu_d$. There are at most $2^B$ endpoint images before duplicate and convex-hull reduction.
\end{proposition}

Holding the other hazards fixed, the demand vector in \eqref{eq:hazard-recursion} is affine, or linear plus a constant, in the remaining hazard. Thus every interior distribution is a weighted average of endpoint distributions (Section~\ref{ec-sec:hazard-endpoints}). Allocations establish compatibility, direct constraints simplify the variables, and endpoints enable finite evaluation.

\subsection{Combining the identified marginals}\label{sec:rectangular}

Carryover requires demand laws for the whole horizon. A \emph{rectangular} family allows any admissible marginal in one period to be paired with any in another. Write $Q$ for the joint law of $O_{1:H}=(O_1,\ldots,O_H)$ and $Q_{1:H}=(Q_1,\ldots,Q_H)$ for the retained marginals. The family is

\begin{equation}
\Pset^{\mathrm{rect}}_\Gamma(Q_{1:H})=\prod_{t=1}^H\Pset_{\Gamma,t}(Q_t).
\label{eq:rect}
\end{equation}

Each point of \eqref{eq:rect} specifies one product demand law for the entire horizon. The law is not selected again after each inventory decision. By Corollary~\ref{cor:recorded-boundary}, retaining the full periodwise laws $Q_t^C$ gives the same family. Joint observation information can impose additional restrictions. Let $\Pset^{\mathrm{full}}_\Gamma(Q)$ contain the product demand laws generated by joint historical mechanisms reproducing the entire $Q$ and satisfying the same periodwise restrictions.

An outer set contains every compatible law and may include additional laws.

\begin{proposition}[Rectangular outer set]\label{prop:rectangular-scope}
Under product demand, $\Pset^{\mathrm{full}}_\Gamma(Q)\subseteq\Pset^{\mathrm{rect}}_\Gamma(Q_{1:H})$, and inclusion can be strict. If admissible historical mechanisms factor across periods and $Q=\bigotimes_tQ_t$, equality holds.
\end{proposition}

Thus the rectangle is exact for retained periodwise information and conservative when joint or stock-flow restrictions narrow the historical class (Section~\ref{ec-sec:full}).

\subsection{One common policy and rectangular data debt}\label{sec:dynamic}

Several demand laws remain possible after identification, so a multistage policy must be evaluated across them. We compare its cost with the informed optimum under the \emph{same} law and take the largest gap across the rectangle. Rectangular data debt is the smallest such gap achievable by one common policy.

\begin{definition}[Rectangular data debt]\label{def:debt}
The deterministic-Markov rectangular data debt is

\begin{equation}
\Delta_{\mathrm{rect}}^M(\Gamma)=\min_{\pi\in\PiM}\sup_{P\in\Pset^{\mathrm{rect}}_\Gamma(Q_{1:H})}\Reg_P(\pi).
\label{eq:debt}
\end{equation}

A minimizer is a rectangular $\Gamma$-robust policy.
\end{definition}

To interpret debt as an information limit, we must show that the rectangle can reproduce the retained logs.

\begin{proposition}[Deterministic reduced-information minimax value]\label{prop:reduced}
Within the historical observation model of Section~\ref{sec:historical-observation}, fix $Q_{1:H}$ and $\Gamma$. A deterministic rule observes only these marginals and returns a policy in $\PiM$. Allow every joint historical mechanism with product demand marginal, observation marginals $Q_t$, and valid periodwise $\Gamma$ restrictions. The smallest attainable worst-case regret is exactly $\Delta_{\mathrm{rect}}^M(\Gamma)$. The same value obtains if the rule instead observes all periodwise full-record laws $Q_t^C$ and mechanisms must reproduce them.
\end{proposition}

Compatible periodwise completions can be multiplied within this observation model to attain every point of \eqref{eq:rect}. Marginalization proves the reverse inclusion. Corollary~\ref{cor:recorded-boundary} extends the construction to fixed $Q_t^C$. This proves both minimax characterizations (Section~\ref{ec-sec:reduced}).

A vertex is a demand law that is not a mixture of others in its set; a product extreme model combines periodwise vertices. These models determine when debt vanishes.

\begin{proposition}[Monotonicity and decision sufficiency]\label{prop:debt}
With the observation law, costs, and policy class fixed, $\Delta_{\mathrm{rect}}^M(\Gamma)$ is nondecreasing in $\Gamma$. It is zero if and only if all product extreme models share an optimal policy in $\PiM$.
\end{proposition}

Thus demand can remain unidentified while the optimal decision is identified. A rule that stays selected as $\Gamma$ varies can still incur positive regret. Section~\ref{ec-sec:common-optimum} proves the zero-debt characterization.

\subsection{From isolated actions to a multistage policy}

A natural shortcut would be to decide whether each period's information identifies its optimal order and then combine those answers. Independent demand does not justify this shortcut, because leftover stock couples the costs of successive decisions. Let $\delta_t$ denote the minimax regret of period $t$ solved alone, from zero inventory and without continuation value. The following two-period family isolates this carryover effect under identical costs.

\begin{proposition}[Carryover under identical period costs]\label{prop:homogeneous}
Let $H=2$, $B=1$, $X_1=0$, and independent demand probabilities $\theta_t=\Prb(D_t=1)\in[\alpha,\beta]$, where $0<\alpha<\beta<1$. In both periods take $K_t=h_t=0$, $b_t=1$, and $c_t=c>\beta$. Then $\delta_1=\delta_2=0$, while

\begin{equation}
\Delta_{\mathrm{rect}}^M(c)=\min\{(2\beta-\beta^2-c)^+,\ (c-2\alpha+\alpha^2)^+\}.
\label{eq:homogeneous}
\end{equation}

It is positive whenever $\max\{\beta,2\alpha-\alpha^2\}<c<2\beta-\beta^2$, a nonempty interval.
\end{proposition}

In isolation, every compatible model prefers no order because $c>\beta$. The same dominance removes second-period ordering from zero in the dynamic problem. The remaining choices are never to order, costing $\theta_1+\theta_2$, or buy one unit initially and never replenish, costing $c+\theta_1\theta_2$. Early stock is worth $\theta_1+\theta_2-\theta_1\theta_2$, the probability of demand occurring at least once. Its range can straddle $c$ even though neither single-period probability does. Section~\ref{ec-sec:homogeneous} proves the reduction.

\begin{figure}[t]
\FIGURE
{\includegraphics[width=.97\linewidth]{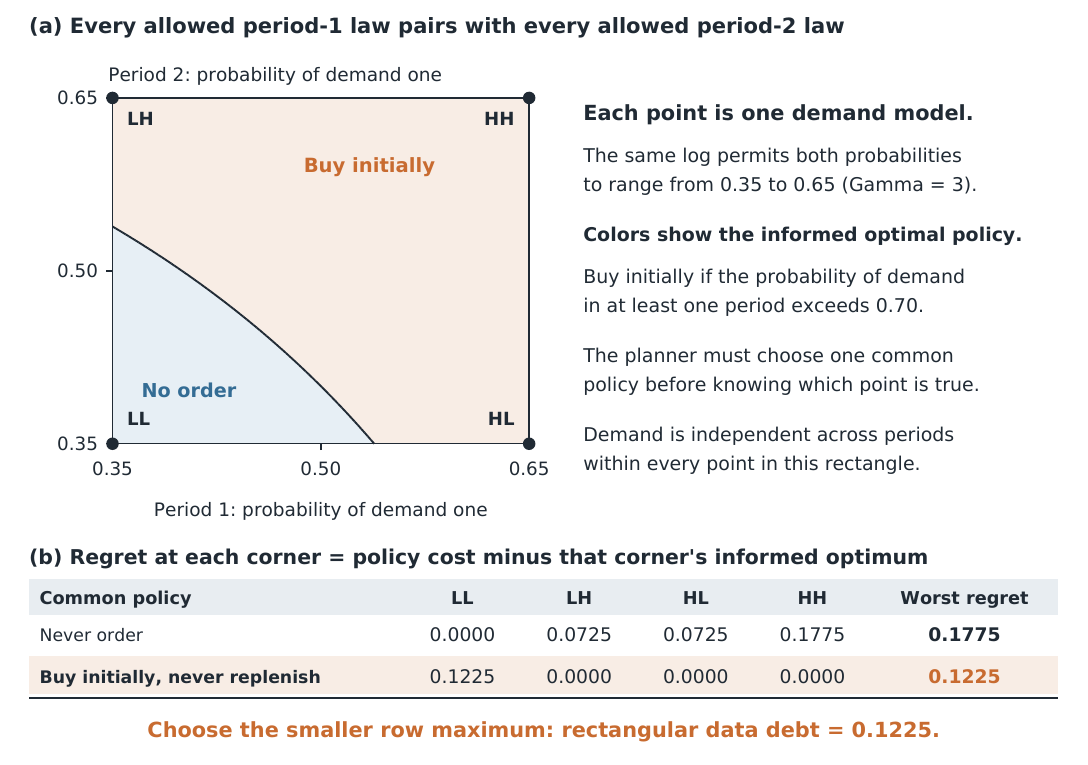}}
{Rectangular data debt in the two-period instance of Proposition~\ref{prop:homogeneous}. (a) Each point is a product demand law with demand-one probabilities $(\theta_1,\theta_2)$. Colors show the informed optimal policy, with indifference on $\theta_1+\theta_2-\theta_1\theta_2=0.70$. L and H denote the endpoints 0.35 and 0.65, in period order. (b) Both undominated deterministic policies are evaluated at the four corners, where their worst regrets occur. The smaller row maximum is $\Delta_{\mathrm{rect}}^M=0.1225$.\label{fig:rectangular-debt}}
{}
\end{figure}

Figure~\ref{fig:rectangular-debt} gives a periodwise censoring instance. Here $o_0=o_1=1/5$, $s_0=3/5$, $s_1=0$, and $\Gamma=3$ yield $\theta_t\in[7/20,13/20]$ in both periods. At $c=7/10$, two zero isolated debts yield deterministic dynamic debt $49/400=0.1225$, attained by initial stocking without replenishment.

Within the stated observation model, the endpoints share identical periodwise sales-and-boundary laws, so deterministic debt persists with exact knowledge of those laws. A deployment mixture that initially stocks with probability $71/120$ has minimax regret $3479/48000\approx0.07248$, which remains positive (Remark~\ref{ec-rem:mixture-benchmark}). With exact log laws, the mixture meets a two-period regret tolerance of 0.10; every deterministic rule has worst-case regret at least 0.1225. The deterministic separation also persists for sufficiently small positive common holding costs (Section~\ref{ec-sec:holding-extension}).

\begin{remark}[Dynamic debt need not add across periods]\label{rem:nonadditivity}
Dynamic debt can also fall below the sum of isolated debts. With time-varying costs, free early stock can remove a later replenishment choice and its positive isolated debt (Section~\ref{ec-sec:reverse-nonadditivity}). Together with Proposition~\ref{prop:homogeneous}, this establishes that there is no general ordering.
\end{remark}

\section{Policy Guarantees and Computation}\label{sec:guarantees}

Population debt assumes known log probabilities. With finite records, a confidence set enlarges the compatible demand set to allow for sampling error. Optimizing over it gives a regret certificate: an upper bound on the selected rule's worst-case regret. We first establish its validity, then bound its gap above debt, derive the sample requirements, and explain computation.

\subsection{From finite records to a policy guarantee}\label{sec:finite-sample}

We observe $N$ iid historical trajectories of $H$ periods. Informative dependence between demand and reveal depth within each trajectory remains allowed. The statistics use $(Y_t,\Delta_t)$, with $\Delta_t$ derived from the recorded boundary. Corollary~\ref{cor:recorded-boundary} justifies the population target; finite-sample protection comes from the confidence construction below.

Hidden selection prevents treating the revealed demand as a representative population sample. It does not prevent estimating the hazard \emph{within} the revealed group. For record $i\in\{1,\ldots,N\}$ at period $t$ and threshold $d<B$, let $A_{i,t,d}=\{Y_{i,t}>d\}\cup\{Y_{i,t}=d,\Delta_{i,t}=1\}$ and $E_{i,t,d}=\{Y_{i,t}=d,\Delta_{i,t}=1\}$. Their counts are $R_{t,d}=\sum_i\ind\{A_{i,t,d}\}$ and $n_{t,d}=\sum_i\ind\{E_{i,t,d}\}$. The first counts records distinguishing stopping from continuation. The second counts stops. Conditioning on that observable risk count gives the following inference target.

\begin{lemma}[Conditional binomial representation]\label{lem:binomial}
For positive population revealed-risk mass, under arbitrary within-period dependence of $(D_t,C_t)$,

\begin{equation}
n_{t,d}\mid R_{t,d}\sim\operatorname{Binomial}(R_{t,d},h^R_{t,d}),\qquad
h^R_{t,d}=\Prb(D_t=d\mid D_t\ge d,C_t>d).
\label{eq:binomial}
\end{equation}

\end{lemma}

The result follows by counting three mutually exclusive cases: revealed demand stopping at $d$, revealed demand exceeding $d$, and all other records (Section~\ref{ec-sec:counting}). It supplies inference for the revealed group's hazard. The sensitivity constraints then relate that hazard to the unrevealed group. A zero observed risk count receives the full interval $[0,1]$. When population risk is zero, the cross-multiplied restrictions handle the boundary directly.

Fix a desired failure probability $\delta\in(0,1)$, distinct from the isolated debts $\delta_t$. The CP intervals bound observable probabilities and revealed hazards \citep{clopper1934}. Dividing $\delta$ across $H(4B+2)$ intervals (Bonferroni allocation) protects their simultaneous coverage. Write $[\underline h_{t,d},\overline h_{t,d}]$ for the interval for $h^R_{t,d}$. Varying the observable probabilities over their intervals and relaxing each sensitivity constraint outward gives the demand projection $\widehat\Pset^{\mathrm{CP}}_{\Gamma,N,t}$. In particular, the unknown true revealed hazard is replaced by the appropriate confidence endpoint in each inequality:

\begin{equation}
\underline h_{t,d}v_{t,d}\le\Gamma(1-\underline h_{t,d})u_{t,d},\qquad
(1-\overline h_{t,d})u_{t,d}\le\Gamma\overline h_{t,d}v_{t,d}.
\label{eq:finite-outer}
\end{equation}

Let $P^0=\bigotimes_{t=1}^H P_t^0$ be the true demand law shared by history and deployment under A5. Section~\ref{ec-sec:outer-confidence} gives the full CP system. CP and the hazard construction in Section~\ref{sec:gamma-sample} are alternative confidence enlargements of the same population set. CP supports the linear implementation, while the hazard recursion exposes the explicit selection-dependent rate.

\begin{theorem}[Containment and a policy certificate]\label{thm:pac}
Under Assumptions~\ref{ass:A1}--\ref{ass:A5} and the independent sampling scheme above, with probability at least $1-\delta$,

\begin{equation}
\Pset_\Gamma^{\mathrm{rect}}(Q_{1:H})\subseteq
\widehat\Pset_{\Gamma,N}^{\mathrm{rect,CP}}:=\prod_t\widehat\Pset^{\mathrm{CP}}_{\Gamma,N,t}.
\label{eq:containment}
\end{equation}

If $\widehat\pi^{\mathrm{CP}}_{\Gamma,N}$ minimizes worst-case regret over this outer set, with value $\widehat\Delta^{\mathrm{CP}}_{\Gamma,N}$, then on the same event,

\begin{equation}
\Reg_{P^0}(\widehat\pi^{\mathrm{CP}}_{\Gamma,N})\le\widehat\Delta^{\mathrm{CP}}_{\Gamma,N},\qquad
\Delta_{\mathrm{rect}}^M(\Gamma)\le\widehat\Delta^{\mathrm{CP}}_{\Gamma,N}.
\label{eq:certificate}
\end{equation}

\end{theorem}

Simultaneous containment precedes policy selection, so it protects the rule chosen from the same data. For any policy define its population worst-case regret by
\begin{equation}
R_{\mathrm{pop}}(\pi)=\sup_{P\in\Pset^{\mathrm{rect}}_\Gamma(Q_{1:H})}\Reg_P(\pi).
\label{eq:selected-pop-risk}
\end{equation}
For a selected rule $\pi$, let $U$ be its checked upper certificate. On the containment event, the benchmark debt, selected-rule risk, certificate, and true-law regret satisfy
\[
\Delta_{\mathrm{rect}}^M(\Gamma)\le R_{\mathrm{pop}}(\pi)\le U,
\qquad \Reg_{P^0}(\pi)\le R_{\mathrm{pop}}(\pi).
\]
Figure~\ref{fig:pac-computation}(b) separates the gap $R_{\mathrm{pop}}(\pi)-\Delta_{\mathrm{rect}}^M$ (policy suboptimality) and $U-R_{\mathrm{pop}}(\pi)$ (assessment conservatism). True-law regret has no general ordering with population debt. Theorem~\ref{thm:pac} requires no uniform clearance, and a rule can be certified before optimization converges (Section~\ref{ec-sec:numerics}).

\subsection{Separating sampling uncertainty from persistent debt}

Containment establishes validity. To assess precision, we measure how far the enlarged set extends beyond the population set, then translate this probability error into cost error. Let $\mathsf I$ denote a fixed confidence construction and $\widehat\Delta^{\mathsf I}_{\Gamma,N}$ its optimized worst-case regret. The largest distance of an outer demand vector from its nearest population-compatible vector, measured by the sum of absolute probability differences, is

\begin{equation}
\mathfrak d_t^{\mathsf I}=
\sup_{\widehat p_t\in\widehat\Pset^{\mathsf I}_{\Gamma,N,t}}
\inf_{p_t\in\Pset_{\Gamma,t}}\|\widehat p_t-p_t\|_1.
\label{eq:distance}
\end{equation}

To translate this error into cost, let $G_t$ be the range of $\widetilde g_t$ over feasible states, actions, and demands, and $S_t$ its largest demand range at a fixed state and action. The clearance floor $\underline\eta_t=\inf_{P_t}P_t(D_t=B)$ is taken over the relevant periodwise family. Demand $B$ clears every feasible stock position. Section~\ref{ec-sec:value} bounds the remaining transformed cost range across starting inventories by $L_t$ and constructs the conversion coefficients $M_t$ as

\begin{equation}
L_{H+1}=0,\quad L_t=G_t+(1-\underline\eta_t)L_{t+1},\quad
M_t=(S_t+L_{t+1})/2,
\label{eq:span-coefficients}
\end{equation}

In Lemma~\ref{lem:perturb}, $p_t,q_t$ are the marginals of two product demand laws $P,Q$; here $Q$ denotes a demand law.

\begin{lemma}[Policy-value perturbation]\label{lem:perturb}
For product laws $P,Q$ in the family defining $M_t$ and any $\pi\in\PiM$,

\begin{equation}
|\widetilde J_P(\pi)-\widetilde J_Q(\pi)|\le\sum_tM_t\|p_t-q_t\|_1.
\label{eq:perturb}
\end{equation}

The same bound holds for the model-specific optimal transformed values.
\end{lemma}

The lemma bounds both a fixed rule's cost and the informed optimum. Regret is their difference, giving the factor two in the certificate gap below. Sections~\ref{ec-sec:value}--\ref{ec-sec:contraction} give the proofs.

\begin{theorem}[Certificate contraction]\label{thm:contraction}
On the simultaneous containment event for the fixed construction $\mathsf I$, with coefficients uniform over its outer family,

\begin{equation}
0\le\widehat\Delta^{\mathsf I}_{\Gamma,N}-\Delta_{\mathrm{rect}}^M(\Gamma)
\le 2\sum_{t=1}^HM_t\mathfrak d_t^{\mathsf I}.
\label{eq:contraction}
\end{equation}

\end{theorem}

\subsection{The sample cost of hidden selection}\label{sec:gamma-sample}

Theorem~\ref{thm:contraction} reduces learning to demand-set error. We next bound this error using sample size and $\Gamma$, then test whether any learner can improve its order. Making the earlier population risk's dependence on $Q_{1:H}$ explicit, write
\begin{equation}
\begin{aligned}
R_{\Gamma,Q}(\pi)&=\sup_{P\in\Pset^{\mathrm{rect}}_\Gamma(Q_{1:H})}\Reg_P(\pi),\\
\mathcal E_{\Gamma,Q}(\pi)&=R_{\Gamma,Q}(\pi)-\Delta_{\mathrm{rect}}^M(\Gamma).
\end{aligned}
\label{eq:gamma-excess-target}
\end{equation}
Thus $\mathcal E_{\Gamma,Q}$ is the selected rule's excess over the best common rule with known log laws. Learning can remove this excess while population debt remains positive. The deployed policy belongs to $\PiM$. The learning algorithm may randomize which such policy it returns.

\paragraph{A confidence set built from the hazards.}
For a revealed stopping probability $u$, define
\[
f_\Gamma(u)=\frac{u}{u+\Gamma(1-u)},\qquad
g_\Gamma(u)=\frac{\Gamma u}{1-u+\Gamma u}.
\]
These hidden-hazard endpoints in \eqref{eq:hazard-recursion} let us track how $\Gamma$ transforms estimation error. Give each observable cell probability a Hoeffding interval and each revealed hazard a conditional binomial Kullback--Leibler (KL) interval $[h_L,h_U]$. Let $\beta=\log(4H(3B+2)/\delta)$. The cell intervals have radius $\sqrt{\beta/(2N)}$, and the hazard interval consists of $u$ satisfying $R_{t,d}\kl(\widehat h_{t,d},u)\le\beta$, where $\widehat h_{t,d}=n_{t,d}/R_{t,d}$ for $R_{t,d}>0$ and $\kl$ denotes binary Kullback--Leibler divergence. At a zero count it is $[0,1]$. Vary the cell vector over its intervals and the probability simplex (nonnegative entries summing to one), vary each hidden hazard over $[f_\Gamma(h_L),g_\Gamma(h_U)]$, and generate demand vectors using \eqref{eq:hazard-recursion}. Their convex hull is $\widehat\Pset^{\mathrm{haz}}_{\Gamma,N,t}$, which contains the population set on simultaneous coverage.

Let $\widehat\pi^{\mathrm{haz}}_{\Gamma,N}$ minimize worst-case regret over the product of these sets, with optimized value $\widehat\Delta^{\mathrm{haz}}_{\Gamma,N}$. A known valid clearance restriction $P_t(D_t=B)\ge\eta$ may be imposed on the outer sets if it holds for every population-compatible marginal. This ensures that sufficiently large demand can erase inventory differences between policies. Section~\ref{ec-sec:gamma-rates} gives the complete construction and proofs.

Here $q$ bounds the joint revealed-risk probability $\Prb(D_t\ge d,C_t>d)$, which includes demand survival as well as stocking depth.

A Hoffman constant converts violations of linear constraints into distance from their feasible set; the hazard construction bounds demand-set error directly.

\begin{theorem}[An explicit hidden-selection upper bound]\label{thm:gamma-upper}
Under Assumptions~\ref{ass:A1}--\ref{ass:A5}, suppose the historical trajectories are iid and every positive revealed-risk probability is at least $q>0$. Suppose $B\ge1$, $\Gamma\ge1$, $G_t\le G$, $S_t\le S$, and the outer families on the simultaneous coverage event have clearance at least $\eta>0$. Set $A=S+G/\eta$ and $\beta=\log(4H(3B+2)/\delta)$. If $Nq\ge8\log(2HB/\delta)$, exact optimization gives, with probability at least $1-\delta$,
\begin{equation}
\begin{aligned}
0\le\mathcal E_{\Gamma,Q}(\widehat\pi^{\mathrm{haz}}_{\Gamma,N})
&\le\widehat\Delta^{\mathrm{haz}}_{\Gamma,N}-\Delta_{\mathrm{rect}}^M(\Gamma)\\
&\le AH\bigl[2\sqrt2(B+1)+8\sqrt2B\sqrt\Gamma\bigr]
\sqrt{\frac\beta{Nq}}.
\end{aligned}
\label{eq:gamma-upper}
\end{equation}
The constants $B,G,S,\eta$ are held fixed when comparing rates in $H,N,q,\Gamma$. No Hoffman constant is required.
\end{theorem}

The odds transforms amplify error measured through square roots of probabilities by at most $2\sqrt\Gamma$, yielding the linear $\Gamma$ sample dependence. Following unresolved mass through successive thresholds converts hazard error into demand-set error. Under shared demands and a common policy, an initial inventory difference then has expected duration at most $1/\eta$, keeping $A=S+G/\eta$ fixed when the cost and clearance bounds are fixed. Section~\ref{ec-sec:gamma-upper-proof} derives outer clearance from population clearance after sufficiently many samples. Without a uniform floor, the present argument yields $H^2\sqrt{\Gamma/(Nq)}$, up to logarithms. The rate describes exact optimization over the hazard sets; CP containment and directional approximation are analyzed separately.

The next result tests necessity using similar logs that require different robust decisions.

\begin{theorem}[Hidden-selection lower bound above positive debt]\label{thm:gamma-lower}
For every $H\ge2$, $\Gamma\ge2$, $0<q\le1/2$, and integer $N\ge1$, there is a family $\{Q_v\}$ of valid binary-demand logging experiments with positive deterministic population debt, revealed risk at least $q$, uniformly bounded known stage costs, and population clearance at least $9/20$. For every learning rule $\mathcal A$ returning a policy in $\PiM$, allowing randomization in that choice,
\begin{equation}
\inf_{\mathcal A}\sup_v\E_v\!\left[\mathcal E_{\Gamma,Q_v}(\mathcal A(\mathrm{data}))\right]
\ge\frac H{2000}\min\left\{1,\sqrt{\frac\Gamma{Nq}}\right\}.
\label{eq:gamma-lower-mean}
\end{equation}
The same family satisfies
\begin{equation}
\inf_{\mathcal A}\sup_v\Prb_v\!\left(
\mathcal E_{\Gamma,Q_v}(\mathcal A(\mathrm{data}))
\ge\frac H{5000}\min\left\{1,\sqrt{\frac\Gamma{Nq}}\right\}\right)\ge\frac15.
\label{eq:gamma-lower-prob}
\end{equation}
Costs are common to the alternatives at fixed $\Gamma,q$ and may vary across these parameter-indexed families. The benchmark is deterministic population debt, not the value for ex ante deployment mixtures.
\end{theorem}

Near revealed hazard $1/(\Gamma+1)$, statistically close logs at fixed revealed risk $q$ require opposite robust actions, each wrong action incurring excess above its own positive debt. Reset periods with certain stock-clearing demand remove carryover between decisions. Adding their losses supplies the factor $H$ in total excess regret.

Comparing both bounds gives sample requirements for the same excess-over-debt target.

\begin{corollary}[Sample complexity with explicit sensitivity]\label{cor:gamma-sample}
Under Theorem~\ref{thm:gamma-upper}, a sufficient trajectory count for $\mathcal E_{\Gamma,Q}(\widehat\pi^{\mathrm{haz}}_{\Gamma,N})\le\varepsilon$ with probability at least $1-\delta$ is
\begin{equation}
N\ge\max\left\{
\frac8q\log\frac{2HB}{\delta},\quad
\frac{288 A^2B^2\Gamma H^2}{q\varepsilon^2}
\log\frac{4H(3B+2)}\delta\right\}.
\label{eq:gamma-samples-upper}
\end{equation}
For $\Gamma\ge2$, $0<q\le1/2$, fixed failure probability $\delta<1/5$, and $0<\varepsilon<H/5000$, Theorem~\ref{thm:gamma-lower} implies the necessary order
\begin{equation}
N=\Omega\!\left(\frac{\Gamma H^2}{q\varepsilon^2}\right).
\label{eq:gamma-samples-lower}
\end{equation}
Thus the $\Gamma,H,q,\varepsilon$ exponents match up to logarithms for the excess-over-debt target, with support, cost and clearance constants fixed. The lower family satisfies these conditions even though its debt is positive.
\end{corollary}

Here $\varepsilon$ bounds total excess regret over $H$ periods. For an average tolerance $\bar\varepsilon=\varepsilon/H$, the upper-bound assumptions and Corollary~\ref{cor:gamma-sample} give the sufficient count
\[
N\ge\max\left\{\frac8q\log\frac{2HB}{\delta},\ \frac{288A^2B^2\Gamma}{q\bar\varepsilon^2}\log\frac{4H(3B+2)}\delta\right\}.
\]
The necessary order is $\Omega(\Gamma/(q\bar\varepsilon^2))$ in the corollary's lower-bound regime, with $0<\bar\varepsilon<1/5000$. Thus the polynomial $H^2$ factor concerns fixed total tolerance; average tolerance retains only logarithmic horizon dependence in the upper bound. The revealed-risk gate still applies. Each of the $N$ trajectories contains $H$ period records.

\paragraph{Relation to noninformative logging.}
Counting each period observation as one scalar record gives $M=HN$ records; the matched order is $\Gamma H^3/(q\varepsilon^2)$ up to logarithms. \citet{qin2026information} already obtain the cubic horizon order under their logging and coverage conditions. Our addition is the explicit selection dependence for learning above positive population debt. Their usable fraction and our joint revealed risk need not coincide. A tolerance below $\Delta_{\mathrm{rect}}^M(\Gamma)$ cannot be met by more records of the same type within $\PiM$.

\subsection{Policy evaluation and numerical certification}\label{sec:scalable}

The statistical results assume exact optimization. Implementation must distinguish selecting a policy from certifying its regret. Each period has $(B+1)!$ feasible lookup tables, so $|\PiM|=((B+1)!)^H$. Proposition~\ref{ec-prop:vertex-app} reduces the search for worst-case demand to at most $2^{HB}$ endpoint products, because a fixed-policy cost difference is affine in each marginal. Small problems permit enumeration. 

Column-and-constraint generation (CCG) solves a smaller policy problem over selected demand scenarios, called the master. It then searches the full set for the chosen policy's worst-case scenario, the separation step, and adds that scenario. A globally solved master gives a lower bound, and complete separation gives a policy upper bound; their gap measures optimization error. Each positive gap adds a new endpoint scenario, ensuring finite termination under exact global solves, but no polynomial-time guarantee. Section~\ref{ec-sec:computation} compares these calculations at the reported small scales.

A finite-sample CP set also varies the observable probabilities. Substituting estimates into the fixed-observation endpoint formula would omit this uncertainty. For each chosen weight vector, or support direction, we bound the weighted sum of demand probabilities. These inequalities give an enclosing polytope $Q_{t,N,K}$ with $K$ directions, distinct from setup cost $K_t$. On coverage, $\Pset_{\Gamma,t}\subseteq\widehat\Pset^{\mathrm{CP}}_{\Gamma,N,t}\subseteq Q_{t,N,K}$. More data address sampling uncertainty. More directions refine the computational enclosure. Section~\ref{ec-sec:contraction} bounds both value errors by the same perturbation argument. Fixed directions can leave a positive gap even with unlimited data.

For $B\le2$, the checked implementation encloses the CP set in a verified polytope. At products of its vertices, dynamic programming (DP) computes upper policy-cost bounds and lower informed-cost bounds, accounting for rounding and vertex-storage error (Section~\ref{ec-sec:numerics}). The largest resulting cost gap gives the selected rule's upper bound $U_{\Gamma,N,K}$. Numerical optimization error, sampling uncertainty, and directional approximation are reported separately.

\section{Numerical Experiments}\label{sec:evidence}

Three studies follow the theoretical sequence: compare finite-log guarantees with population debt, vary costs to locate order changes, and test common optimality after deeper revelation. All policies are deterministic and use the same-law informed comparator.

\subsection{Policy learning and guarantees from finite logs}\label{sec:controlled}

\paragraph{Comparing the confidence constructions.}
We first use binary demand, taking values zero or one, to compare policy risk and bound tightness for the hazard construction of Theorem~\ref{thm:gamma-upper} and the specialized five-interval CP construction on identical logs. We use $B=1,H=2$, $\Gamma\in\{2,5,10\}$, $q\in\{0.1,0.3\}$, and both signs of a fixed hazard perturbation around the lower-bound decision threshold. Each of these 12 instances keeps its demand law and costs fixed across $N=250,1{,}000,4{,}000$, with 200 paired datasets per cell and $\delta=0.05$. Four policies and all endpoint products are enumerated. Section~\ref{ec-sec:hazard-cp} gives construction and coverage details.

Table~\ref{tab:hazard-cp} averages the fixed instances. Specialized CP has lower mean interval width and certificate in all 36 design cells. Each paired 95\% interval, computed separately for a fixed design cell, excludes zero. Mean excess population risk favors hazard in eight cells, CP in nine, and ties in 19. The paired intervals exclude zero in three hazard-favoring cells and four CP-favoring cells. In the negative-perturbation instances, the true-law oracle differs from the population minimax policy. Learning the latter can therefore increase true-law regret. The constructions differ in intervals, parameterization and coverage allocation; this fixed-design comparison does not isolate interval efficiency or establish an asymptotic rate.

\begin{table}[t]
\centering
\begin{threeparttable}
\caption{Paired binary confidence constructions: design-averaged results}
\label{tab:hazard-cp}
\small
\begin{tabular*}{\textwidth}{@{\extracolsep{\fill}}rlrrrr@{}}
\toprule
$N$ & Method & $U$ & Excess risk & True-law regret & Width \\
\midrule
250 & Hazard & 0.2547 & 0.001849 & 0.002936 & 0.7559 \\
250 & CP (binary) & 0.2370 & 0.001850 & 0.002998 & 0.7038 \\
1,000 & Hazard & 0.2299 & 0.001809 & 0.007410 & 0.5962 \\
1,000 & CP (binary) & 0.2102 & 0.001828 & 0.005479 & 0.5493 \\
4,000 & Hazard & 0.2068 & 0.001798 & 0.010757 & 0.4808 \\
4,000 & CP (binary) & 0.1904 & 0.001784 & 0.007320 & 0.4484 \\
\bottomrule
\end{tabular*}
\begin{tablenotes}\footnotesize
\item \textit{Notes.} Each entry equally weights 12 fixed instances and 200 paired datasets per instance. CP uses the specialized five-interval construction. Excess risk is $R_{\rm pop}-\Delta_{\rm rect}^M$; width refers to the first-period zero-demand probability. Mean debt is 0.165034. These design averages do not pool the instances into an iid confidence calculation.
\end{tablenotes}
\end{threeparttable}
\end{table}

\begin{figure}[t]
\FIGURE
{\includegraphics[width=\textwidth]{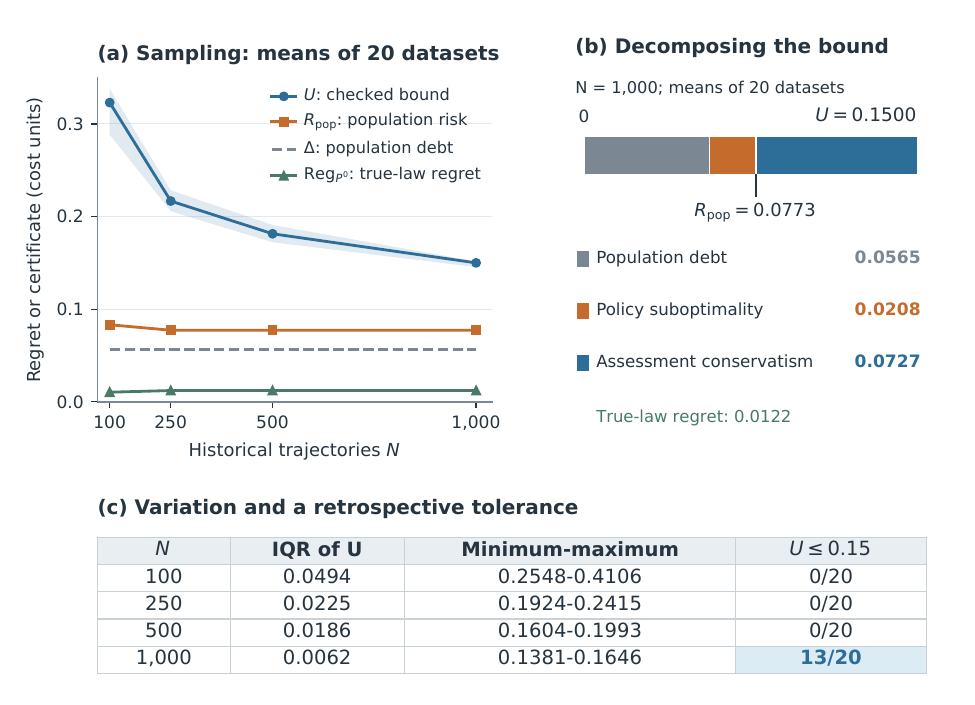}}
{Controlled CP diagnostics at $\Gamma=5.5$. (a) Means over 20 datasets. Shading is the interquartile range of $U$. (b) Population debt, selected-policy suboptimality, and assessment conservatism at $N=1{,}000$. (c) Dispersion and the count meeting tolerance 0.15.\label{fig:pac-computation}}
{}
\end{figure}

\paragraph{Policy error and assessment conservatism.}
The next study adds a third demand value and separates policy error above debt from conservatism in the bound. It uses $H=3$, true and candidate $\Gamma=5.5$, eight support directions, and 20 datasets per $N$. From $N=250$ to 1,000, population risk remains 0.0773 while the mean checked bound falls from 0.2167 to 0.1500. More records chiefly tighten the guarantee over this range. At $N=1{,}000$, true-law regret is 0.0122 and debt is 0.0565. Policy suboptimality is 0.0208 and assessment conservatism is 0.0727, as Figure~\ref{fig:pac-computation}(b) shows. Both sampling and projection can affect selection. Each selected policy is evaluated on all 64 population endpoint products.

The eight-direction population optimum is 0.0678, exceeding debt by 0.0113. On one separate $N=1{,}000$ sample, 64 directions change the bound from 0.151021 to 0.150956. These are not replication means. Figure~\ref{fig:pac-computation}(c) reports dispersion and shows that 13 of 20 bounds meet 0.15 at $N=1{,}000$. In a separate binary check with true sensitivity 3, using $\Gamma=1.5$ lowers demand containment from 98.5\% to 25.4\% as $N$ rises from 100 to 1,000. More data cannot repair a misspecified class (Section~\ref{ec-sec:controlled}).

\subsection{When costs change the decision}\label{sec:semisynthetic}\label{sec:retail}

We vary costs on fixed logs to identify when demand uncertainty changes replenishment. Naturally uncensored two-day FreshRetailNet paths \citep{wang2025} are a nonrandom sample, scaled and clipped to demand proxies in $\{0,1,2\}$ and artificially censored. Training through November 24, 2024 alone determines selection, scales, and capacity calibration. Of 2,000 selected series, 1,177 supply qualifying held-out paths. The formal target pairs independent draws from the periodwise training distributions. Held-out targets stress stability and dependence. These semi-synthetic results do not estimate retailer demand, hidden selection, or realized savings (Section~\ref{ec-sec:retail}).

Plug-in inference treats empirical log frequencies as exact; CP allows them to vary over confidence intervals. We pair each method with minimax regret or worst expected original cost, naming the objectives regret and cost. Original cost retains the demand term that cancels in same-law regret. Adverse censoring lowers the demand-dependent reveal depth as demand rises; aligned censoring raises it. The grid fixes 30 adverse $N=1{,}000$ logs and varies $K\in\{0.1,0.5,1\}$, $h\in\{0,0.2,0.6\}$, and shortage cost $b\in\{0.5,1,2\}$, with $c=0$. Both periods share these costs. The generator's maximum selection odds ratio is 6.9872. Thus $\Gamma=10$ covers it, whereas $\Gamma=1$ is a misspecified benchmark. Every rule is assessed on the same $\Gamma=10$ population family and log-specific CP outer set.

At $\Gamma=10$, CP-regret and plug-in-regret behave identically in all runs in 18 of 27 cost cells. In the other nine, behavior changes in 3.3\%--100\% of runs. Table~\ref{tab:cost-grid-summary} shows mixed risk effects despite lower common certificates, which are themselves the CP selection objective. Table~\ref{ec-tab:cost-grid-all} retains all 27 cells.

At $(K,h,b)=(0.1,0.6,1)$, all 30 CP policies replenish second-period inventory $X_2=1$ to two units. Plug-in keeps one. The extra unit costs $K+h-(h+b)P_2(D_2=2)$ conditional on this state. It raises cost throughout the population family because the tail probability is at most 0.3875, below the break-even level 0.4375. All CP assessment sets admit tails above that threshold. The extra order therefore protects against tails outside the population family. Mean certificate falls by 0.0100 while excess population risk rises by 0.0712 and true-law regret by 0.0344.

At $(1,0.6,2)$, CP instead raises the second-period order from one to two when inventory is empty in 9 of 30 runs. Both actions already pay the fixed fee, so the extra unit costs $h-(h+b)P_2(D_2=2)$. Its break-even tail is 0.2308, below the true value 0.2883. Mean true-law regret falls by 0.0121 and excess population risk by 0.0031. Whether another fixed fee is incurred distinguishes these two responses to uncertainty.

\begin{table}[t]
\centering
\begin{threeparttable}
\caption{Cost conditions separate changes in certificates from changes in risk}
\label{tab:cost-grid-summary}
\small
\begin{tabular*}{\textwidth}{@{\extracolsep{\fill}}lrrrr@{}}
\toprule
Metric & Lower & Equal & Higher & Range of mean differences \\
\midrule
\multicolumn{5}{@{}l}{CP-regret minus plug-in-regret, both at $\Gamma=10$} \\
Common certificate $U$ & 9 (7) & 18 & 0 (0) & $[-0.0522,0.0000]$ \\
Excess population risk & 6 (4) & 18 & 3 (3) & $[-0.0089,+0.0778]$ \\
True-law regret & 6 (4) & 18 & 3 (3) & $[-0.0314,+0.0523]$ \\
\addlinespace[3pt]
\multicolumn{5}{@{}l}{CP-regret: $\Gamma=10$ minus $\Gamma=1$} \\
Common certificate $U$ & 14 (13) & 13 & 0 (0) & $[-0.4223,0.0000]$ \\
Excess population risk & 12 (11) & 13 & 2 (2) & $[-0.0560,+0.0712]$ \\
True-law regret & 11 (11) & 13 & 3 (2) & $[-0.2154,+0.0344]$ \\
\bottomrule
\end{tabular*}
\begin{tablenotes}\footnotesize
\item \textit{Notes.} Counts are among 27 cost cells, each using the same 30 logs. Parentheses count cells whose pointwise paired 95\% $t$ interval excludes zero in that direction; there is no multiplicity adjustment. All policies use the same $\Gamma=10$ population and CP assessment sets. Lower $U$ is the CP-regret selection objective, not by itself an operating gain.
\end{tablenotes}
\end{threeparttable}
\end{table}

Replacing $\Gamma=1$ by 10 in CP-regret lowers mean true-law regret in 11 cost cells and raises it in three. Ignoring hidden selection can therefore be costly, but widening the set need not improve true-law performance everywhere. Changing the CP objective to worst original cost lowers that bound in nine cells while raising regret certificates and excess risk in those cells. The cost parameters determine which uncertainty matters for action and which criterion the policy protects.

At baseline costs, four-method performance agrees at $N=1{,}000$. For adverse $N=250$ logs, the CP-regret population-risk difference is $-0.00614$, with paired interval $[-0.01313,0.00086]$. The mean improvement remains inconclusive. Under aligned censoring, CP-cost raises true-law regret by 0.02763, interval $[0.00088,0.05439]$, while its mean worst-cost-bound change $-0.00087$ has interval $[-0.00192,0.00017]$. All intervals are pointwise and conditional on fixed calibration and path pools.

\paragraph{Evaluation targets and computation.}
Holding high-service-cost policies fixed, we compare plug-in candidates 1 and 10, the two $\Gamma$ values. Evaluation can pool series before drawing independently by period, draw a series first and then its independent period demands, or retain empirical two-day pairs. The corresponding mean cost changes are $-0.0433$ for the pooled marginal product, $+0.0031$ for a mixture of series products, and $+0.0521$ for empirical joint paths. Their single-period marginals coincide, yet rankings reverse. Changing $\Gamma$ cannot repair pooling or temporal dependence. Section~\ref{ec-sec:pooled-dependence} gives paired intervals. Computation covers 36 or 216 complete policies for $B=2,H=2$ or 3. Independent enumeration confirms the stored-set optima. Section~\ref{ec-sec:computation} reports timings for these enumerated problems.

\subsection{A controlled revelation study}\label{sec:pilot}

Deeper revelation supplies information about previously hidden demand. We test whether it establishes common optimality while demand remains partly identified. Two binary periods share a true observation marginal, estimated from one sample of 1,000 marginal observations. Both CP methods reuse this sample and growing prefixes of one independent revelation stream. CP/direct retains an interval for the hidden group's stopping probability; CP/$\Gamma$ compresses it into a symmetric $\Gamma$ bound. At 1,600 new observations, CP/direct has zero bounds in all 500 runs per scenario, while CP/$\Gamma$ retains means 0.3808 and 0.4040. Both methods initially select true-law oracle policies. The zero bounds certify common optimality over demand sets containing multiple distributions, as in Proposition~\ref{prop:debt}, without a positive operating gain. Smaller samples produce 18 cost increases after CP/direct policy updates. Collection costs are not deducted, and the guarantee is for fixed sample size, not choosing when to stop after inspecting the bound (Section~\ref{ec-sec:pilot-hazards}).

\section{Discussion and Conclusion}\label{sec:discussion}

The analyst validates sales, reveal depths, pooling, independence, population stability, and logging architecture before choosing a selection range (Section~\ref{ec-sec:calibration}). An audit records intended stock before independently assigning full revelation. Under demand-preserving revelation and stable intended stocking, audited demand identifies threshold risk groups. Hazard intervals can bound odds or constrain demand directly. Calibrated ranges require calibration-error control.

A rule qualifies when its checked bound meets a cost-unit regret tolerance per horizon or period across the credible selection range. More records address sampling error; refined computation addresses approximation; deeper revelation or richer rules can change debt. Collection uses a prespecified budget with fixed sample sizes or planned checkpoints with simultaneous protection.

Revelation may certify common optimality without identifying demand. Acquisition compares guarantee changes, old and new policy costs under a common law, and collection costs over future uses.

\bibliographystyle{informs2014}
\bibliography{references}

\clearpage
\DDStartCompanion
\pdfbookmark[0]{Electronic Companion}{electronic-companion}
\begin{center}{\Large\bfseries Electronic Companion to The Data Debt of Stockouts\par}\smallskip
{\normalsize Partial Identification and Minimax-Regret Inventory Control}\end{center}
\noindent Sections~\ref{sec:identification}--\ref{sec:gamma-rates} contain proofs; Section~\ref{sec:experiments} gives experimental details.

\begin{table}[htbp]
\centering
\begin{threeparttable}
\caption{Information restrictions and targets in adjacent learning frameworks}
\label{tab:technical-comparison}
\footnotesize
\setlength{\tabcolsep}{4pt}
\renewcommand{\arraystretch}{1.0}
\begin{tabularx}{\textwidth}{@{}>{\RaggedRight\arraybackslash}p{.19\textwidth}*{2}{>{\RaggedRight\arraybackslash}X}@{}}
\toprule
Study & Information restriction & Decision target \\
\midrule
Qin et al.\ (2026) & Demand law fixed conditional on the full boundary array. & Inventory policy gaps and value estimation from scalar records. \\
\addlinespace[2pt]
Han et al.\ (2026) & History-based stocking; current demand law fixed given history. & Offline policy gaps; online cumulative regret. \\
\addlinespace[2pt]
Jin et al.\ (2026) & Marginal sensitivity model for unobserved treatment confounding. & Same-law class-relative minimax regret; randomized deployment allowed. \\
\addlinespace[2pt]
Hess et al.\ (2026) & Marginal sensitivity model for unobserved treatment confounding. & Sharp value bounds, policy learning, and baseline improvement. \\
\addlinespace[2pt]
This paper & Nested sales/stocking records with bounded hidden-selection odds. & Common deterministic rule; learning above debt from iid trajectories. \\
\bottomrule
\end{tabularx}
\begin{tablenotes}\footnotesize
\item \textit{Notes.} The three inventory settings share sales and stocking boundaries as observables. Treatment-policy settings have different records and policy classes. Our revealed-risk probability includes demand survival and is not a stocking-only coverage fraction.
\end{tablenotes}
\end{threeparttable}
\end{table}

\section{Identification and Data Debt}\label{sec:identification}\label{sec:prelim}

Notation follows the manuscript; periodwise arguments suppress $t$. Norms are $\ell_1$ unless specified, and $\dist_1(p,A)=\inf_{a\in A}\|p-a\|_1$. The transformed cost in \eqref{main-eq:transformed-cost} subtracts $b_tD_t$, independently of the policy, so
\begin{equation}
J_P(\pi)-J_P^*=\widetilde J_P(\pi)-\widetilde J_P^*.
\label{eq:regret-transform}
\end{equation}
For physical demand with finite first moments and a fixed action cap $B$, top coding $D_t^B=\min\{D_t^{\rm phys},B\}$ preserves sales and inventory paths. Only shortage cost changes, by $b_t(D_t^{\rm phys}-B)^+$, whose expected sum is common to every policy and its class oracle. Thus regret is unchanged. Changing the action cap changes the policy class and is a separate comparison.

\subsection{One completion for both observation records}\label{sec:latent}

\begin{proof}[Proof of Theorem~\ref{main-thm:sharp} and Corollary~\ref{main-cor:recorded-boundary}]
Compatibility here refers to the historical observation model of Section~\ref{main-sec:historical-observation}. Any compatible joint law supplies the latent masses in \eqref{main-eq:latent}, the marginal projection, and the cross-multiplied restrictions \eqref{main-eq:gamma1}--\eqref{main-eq:gamma2}; necessity follows. Conversely, choose a feasible hidden array $m$ and revealed cells $a_{k,c}\ge0$ for $k<c$ with $\sum_{c>k}a_{k,c}=o_k$. Define
\[
\Prb(D=k,C=c)=m_{k,c}\quad(c\le k),\qquad
\Prb(D=k,C=c)=a_{k,c}\quad(k<c).
\]
For a coarsened record, take $a_{k,B+1}=o_k$ and the other revealed cells zero. For a given full periodwise record $Q^C$, take its observed revealed cells. The array has total mass $\sum_cs_c+\sum_ko_k=1$ and reproduces every required record: revealed cells are fixed by $a$, and censored cell $c$ has mass $\sum_{k\ge c}m_{k,c}=s_c$. Its demand marginal is $p_k=o_k+\sum_{c\le k}m_{k,c}$. The revealed stopping and continuing masses remain $o_d$ and $\Prb(Y>d)=\tau_d$, and the hidden masses depend only on $m$. Thus the same $\Gamma$ restrictions hold for both records. This construction uses no division and includes zero-mass cases; if the latent set is empty, both demand sets are empty.
\end{proof}

\begin{proof}[Proof of Proposition~\ref{main-prop:gamma1}]
At $\Gamma=1$, $e_dv_d=\tau_du_d$. Since $r_d=e_d+\tau_d>0$, this implies
$r_d(e_d+u_d)=e_d(e_d+u_d+\tau_d+v_d)$, so the demand hazard is $e_d/r_d$. These hazards uniquely determine the product-limit distribution, including boundary values zero and one.
\end{proof}

\subsection{A common construction for the two demand representations}\label{sec:direct-projection}\label{sec:hazard-endpoints}

\begin{proof}[Proof of Propositions~\ref{main-prop:direct-demand} and~\ref{main-prop:endpoint-hazard}]
Every feasible allocation satisfies the direct-demand constraints: $p_k-o_k$ is its row sum, columns $c\ge d$ can use only rows $k\ge d$, and the hidden event and tail masses are
\[
u_d=p_d-o_d,\qquad v_d=\sum_{k>d}p_k-\tau_d.
\]
Conversely, take a probability vector satisfying those constraints. They give $u_d,v_d\ge0$. For $d<B$, put $a_d=u_d+v_d$; normalization and probability accounting give $a_0=s_0$ and $a_{d+1}=v_d+s_{d+1}$, with terminal $a_B=p_B-o_B$. If $a_d>0$, set $\nu_d=u_d/a_d$; otherwise choose any point in the hazard interval of \eqref{main-eq:hazard-recursion}. The $\Gamma$ inequalities give exactly that interval, including the stated zero-mass conventions. Consequently the recursion in \eqref{main-eq:hazard-recursion} recovers $p$.

For any hazards in that box, a compatible allocation is
\begin{equation}
m_{k,c}=s_c\nu_k\prod_{j=c}^{k-1}(1-\nu_j)\quad(c\le k<B),\qquad
m_{B,c}=s_c\prod_{j=c}^{B-1}(1-\nu_j).
\label{eq:common-hazard-completion}
\end{equation}
Each column sums to $s_c$. Its unresolved mass at $d$ is $\sum_{c\le d}s_c\prod_{j=c}^{d-1}(1-\nu_j)=a_d$, giving row sum $\nu_da_d=u_d$ and terminal row sum $a_B$. The hazard bounds ensure the sensitivity restrictions. This proves sufficiency of the direct constraints and that the hazard box maps onto the demand set. The direct description has $B+1$ variables and $O(B)$ constraints. Finally, the recursion is affine in each hazard separately, so multilinear interpolation expresses every image as a convex combination of endpoint images. All endpoint images are feasible and the demand set is convex, proving the endpoint representation.
\end{proof}

\subsection{Periodwise information and the common policy}\label{sec:full}\label{sec:reduced}

\begin{proof}[Proof of Propositions~\ref{main-prop:rectangular-scope} and~\ref{main-prop:reduced}]
Marginalizing any compatible historical mechanism gives a member of $\prod_t\Pset_{\Gamma,t}(Q_t)$. Conversely, multiplying compatible periodwise completions attains any point of that rectangle and reproduces the specified $Q_t$, or $Q_t^C$ by Corollary~\ref{main-cor:recorded-boundary}. This product construction is admissible because the historical observation model imposes no interperiod stock-flow constraint. It proves demand-image equality for marginal-only information, even when allowed mechanisms also include dependent records. At a fixed information input a deterministic rule returns one $\pi\in\PiM$; every such output is attainable by a constant rule. Minimizing its worst regret over the demand image therefore gives exactly \eqref{main-eq:debt}.

If the full observation law $Q$ is fixed, marginalization gives inclusion in the rectangle. For $Q=\bigotimes_tQ_t$, the product completions also reproduce $Q$ and prove equality under factorized mechanisms. The following example establishes that inclusion can otherwise be strict.
\end{proof}

\begin{example}[A strict inclusion]\label{ex:strict-rectangular-app}
In the unrestricted reveal-depth model, let $D_1,D_2$ be independent Bernoulli$(1/2)$. Reveal both at depth two when they agree; otherwise censor both at zero. The observation law gives mass $1/4$ to each of the jointly revealed zeros and ones, and $1/2$ to joint censoring. Each period has $o_0=o_1=1/4$, $s_0=1/2$, $s_1=0$, and the generating mechanism satisfies $\Gamma=1$. At candidate $\Gamma=10$, the periodwise interval is $p_t(0)\in[13/44,31/44]$. Its lower-endpoint product violates $p_1(0)p_2(0)\ge1/4$, which is necessary to reproduce the jointly revealed-zero event. Thus the full observation law excludes a point of the rectangle.
\end{example}

\subsection{Endpoint evaluation and zero debt}\label{sec:scenario-master}\label{sec:common-optimum}

\begin{proposition}[Product-vertex reduction]\label{prop:vertex-app}
If $\Pset_t=\operatorname{conv}(\mathcal E_t)$ with finite $\mathcal E_t$, then every $\pi\in\PiM$ satisfies
\[
\sup_{P\in\prod_t\Pset_t}\{J_P(\pi)-J_P^*\}
=\max_{e\in\prod_t\mathcal E_t}\{J_e(\pi)-J_e^*\}.
\]
\end{proposition}
\begin{proof}
Write regret as $\max_{\pi'\in\PiM}\{J_P(\pi)-J_P(\pi')\}$. For a fixed pair, its expectation is the sum of pathwise cost differences weighted by $\prod_tp_t(d_t)$, hence is affine in each marginal separately. Successive maximization therefore attains a product of generating points. The outer supremum commutes with the finite policy maximum, proving the claim.
\end{proof}

\begin{proof}[Proof of Proposition~\ref{main-prop:debt}]
Increasing $\Gamma$ enlarges the demand sets and cannot reduce minimax regret. Finiteness of $\PiM$ guarantees a minimizer. Its nonnegative worst regret is zero exactly when it is optimal at every product extreme model, by Proposition~\ref{prop:vertex-app}.
\end{proof}

\subsection{Carryover and isolated-period debt}\label{sec:homogeneous}\label{sec:holding-extension}\label{sec:reverse-nonadditivity}

\begin{proof}[Proof of Proposition~\ref{main-prop:homogeneous}]
Since $c>\beta$, no order is optimal in each isolated period and dominates replenishment from zero in period two. At inventory one no positive order is feasible. Replacing any period-two rule by no replenishment weakly lowers its cost under every demand law; the same reduction applies to the informed oracle. Only two undominated choices remain: never order, costing $J_A=\theta_1+\theta_2$, or order initially and never replenish, costing $J_B=c+\theta_1\theta_2$. The difference is governed by the increasing function $\theta_1+\theta_2-\theta_1\theta_2$. Thus their worst regrets are $(2\beta-\beta^2-c)^+$ and $(c-2\alpha+\alpha^2)^+$, giving \eqref{main-eq:homogeneous}. The positive-debt interval is nonempty because $2\beta-\beta^2$ exceeds both $\beta$ and $2\alpha-\alpha^2$.
\end{proof}

For the main figure's instance, $[\alpha,\beta]=[7/20,13/20]$ and $c=7/10$ give worst regrets $71/400$ and $49/400$. The isolated and final-period dominance margin is $c-\beta=1/20$ and remains strict with nonnegative holding cost. For each policy, costs are affine in the common holding cost $h$; the policy class is finite and the demand rectangle compact. Worst regrets and their minimum are therefore continuous in $h$. The strict positive-debt separation and initial-order preference persist for sufficiently small $h>0$.

\begin{remark}[Deployment randomization in Figure~\ref{main-fig:rectangular-debt}]\label{rem:mixture-benchmark}
Allow an ex ante mixture, retaining the deterministic oracle. Let $\lambda$ be the probability of initial stocking; otherwise never order. The demand law is fixed before the independent policy draw. The preceding dominance argument removes second-period ordering from every mixture component. Since $J_A-J_B=\theta_1+\theta_2-\theta_1\theta_2-c$ increases over the rectangle, worst expected regret is
\[
R_{\rm mix}(\lambda)=\max\left\{\lambda\frac{49}{400},(1-\lambda)\frac{71}{400}\right\}.
\]
Balancing the terms gives $\lambda^*=71/120$ and $\min_\lambda R_{\rm mix}(\lambda)=3479/48000\approx0.07248$. The mixed regret remains positive despite zero isolated debts.
\end{remark}

\begin{proof}[Reverse example for Remark~\ref{main-rem:nonadditivity}]
Let $H=2$, $B=1$, $X_1=0$, and $D_1=0$ be fully revealed. Set all first-period costs to zero. At $\Gamma=4$, second-period records $o_2=(1/9,4/21)$ and $s_2=(44/63,0)$ give $\theta_2=\Prb(D_2=1)\in[2/5,4/5]$ by \eqref{main-eq:hazard-recursion}. Take $(K_2,c_2,h_2,b_2)=(2/5,1/10,0,4/5)$. In isolation, not ordering costs $4\theta_2/5$ and ordering costs $1/2$, with worst regrets $7/50$ and $9/50$. Thus $\delta_1+\delta_2=7/50$. Buying one unit for free in period one and carrying it forward incurs zero cost under every compatible law. Nonnegative costs imply a zero oracle value as well, hence $\Delta_{\mathrm{rect}}^M=0$.
\end{proof}

\subsection{Information scope and sensitivity calibration}\label{sec:timing}\label{sec:depth-information}\label{sec:calibration}

For any periodwise law of $(D,C)$, draw $S$ with $C$'s marginal, set $C=S$, then draw $D$ conditionally on $C$. This imposes no stock-flow equations. Inventory without removals and with nonnegative replenishment also requires $C_{t+1}\ge C_t-Y_t$. These joint restrictions shrink the rectangle, even with periodwise records; its minimax value bounds the narrower class from above. Recorded-boundary equivalence uses the stated model. The retail proxy supplies calibration; treating it as an effective boundary requires validation.

Certificates retain every row $(t,d)\in\mathcal U$. Externally justified revealed- and hidden-hazard intervals bound both odds ratios; their largest bound across $\mathcal U$ gives $\Gamma^{\rm deploy}$ under the extended boundary convention. These intervals need external support, such as validated deeper revelation; stocking or recorded forecasts alone leave hidden hazards unidentified. Zero observed risk receives $[0,1]$.

\section{Confidence and Numerical Guarantees}\label{sec:finite}

Hats here denote the experimental CP construction. The explicit rate in Section~\ref{sec:gamma-rates} concerns a separate hazard family.

\subsection{Simultaneous coverage and containment}\label{sec:counting}\label{sec:outer-confidence}

Suppress the period index. Form two-sided CP intervals for the $2B+2$ observable cells $o_d,s_c$, the $B$ risk probabilities $r_d$, and the $B$ revealed hazards, allocating error $\delta/[H(4B+2)]$ to each. At $R_d=0$ use $[0,1]$ for the hazard. The latent confidence polytope $\widehat\Zset_{\Gamma,N}$ varies $(o,s)$ over their intervals and the probability simplex, imposes
\[
\sum_{k=c}^B m_{k,c}=s_c,\quad m_{k,c}\ge0,\qquad
p_k=o_k+\sum_{c\le k}m_{k,c},
\]
and requires $r_d=o_d+\sum_{y>d}(o_y+s_y)$ to lie in its interval. For $u_d,v_d$ defined in \eqref{main-eq:hidden-masses}, impose the two outer inequalities \eqref{main-eq:finite-outer}. Its demand projection is $\widehat\Pset^{\rm CP}_{\Gamma,N}$.

\begin{proof}[Proof of Lemma~\ref{main-lem:binomial} and Theorem~\ref{main-thm:pac}]
The counts in $E_d$, $A_d\setminus E_d$, and $A_d^c$ are multinomial. For $r_d>0$, conditioning on the first two counts totaling $R_d$ makes the first count binomial with probability $h_d^R=e_d/r_d$. Conditional CP coverage, averaged over $R_d$, therefore gives marginal coverage for the hazard; cell and risk counts are ordinary binomials. A zero population risk has $R_d=0$ almost surely. The union bound yields an event $\mathcal E_N$ of probability at least $1-\delta$ covering all intervals.

On this event, every population-feasible allocation satisfies, on positive-risk rows,
\[
\begin{aligned}
\underline h_dv_d\le h_d^Rv_d\le\Gamma(1-h_d^R)u_d
&\le\Gamma(1-\underline h_d)u_d,\\
(1-\overline h_d)u_d\le(1-h_d^R)u_d\le\Gamma h_d^Rv_d
&\le\Gamma\overline h_dv_d.
\end{aligned}
\]
Zero-risk rows have $\underline h_d=0,\overline h_d=1$, making both outer inequalities vacuous. True cells satisfy all confidence constraints. Hence the entire population latent set, its demand projection, and their periodwise products are contained in the outer sets. Since $P^0$ belongs to the product set, its regret is bounded by the optimized supremum. Monotonicity under set inclusion gives the population-debt inequality.
\end{proof}

\begin{lemma}[Binary coefficient extrema]\label{lem:binary-coefficient-outer-app}
For $B=1$, put $e=o_0$, $z=o_1+s_1$. Given simultaneous bounds $[e_L,e_U],[z_L,z_U]$ and an outer region $\mathcal R_N$ for $(o_0,s_0)$, set
\[
\underline k=\frac{e_L}{e_L+\Gamma z_U},\qquad
\overline k=\frac{\Gamma e_U}{z_L+\Gamma e_U}.
\]
Use $\underline k=0$ if $e_L=0$ and $\overline k=1$ if $z_L=0$. Then $p_0$ lies between $\min_{\mathcal R_N}(o_0+\underline k s_0)$ and $\max_{\mathcal R_N}(o_0+\overline k s_0)$.
\end{lemma}
\begin{proof}
Writing $x=\Prb(D=0,C=0)$, the restrictions are $e(s_0-x)\le\Gamma zx$ and $zx\le\Gamma e(s_0-x)$. Their solution coefficients increase in $e$ and decrease in $z$. Endpoint extrema and $p_0=o_0+x$ prove the bounds, including the empty-group conventions.
\end{proof}

\subsection{Value spans and perturbation}\label{sec:value}

Write $\spn f=\max f-\min f$ over the indicated finite domain.

\begin{proof}[Proof of Lemma~\ref{main-lem:perturb}]
Use $G_t,S_t,L_t,M_t$ from \eqref{main-eq:span-coefficients}, uniform over the rectangular family under consideration. Let $\widetilde V_{t,P}^{\pi}(x)$ be the remaining transformed cost, with terminal value zero. Couple two starting inventories using the same demands and policy. Current costs differ by at most $G_t$; demand $B$ sends both next states to zero. Backward induction therefore gives
\begin{equation}
\spn_x\widetilde V_{t,P}^{\pi}(x)
\le G_t+(1-p_t(B))\spn_x\widetilde V_{t+1,P}^{\pi}(x)\le L_t.
\label{eq:span-recursion-app}
\end{equation}
Replace $p_t$ by $q_t$ one period at a time. Adjacent hybrids have the same starting-state distribution at that period and the same future model. Conditional on state $x$, the range over $d$ of $\widetilde g_t(x,\pi_t(x),d)+\widetilde V_{t+1}^{\pi}(\pi_t(x)-\min\{d,\pi_t(x)\})$ is at most $S_t+L_{t+1}$. Total variation bounds the expectation difference by $M_t\|p_t-q_t\|_1$. Summing gives the policy bound; taking minima gives the oracle bound because the coefficients are uniform over policies.
\end{proof}

\subsection{A common set-distance bound}\label{sec:contraction}

\begin{proof}[Proof of Theorem~\ref{main-thm:contraction}]
For nonempty periodwise sets $\mathcal A_t,\mathcal B_t$, let $d_t=\sup_{a\in\mathcal A_t}\inf_{b\in\mathcal B_t}\|a-b\|_1$, with $M_t$ uniform over $\prod_t(\mathcal A_t\cup\mathcal B_t)$. The preceding lemma bounds the policy and oracle changes separately. Choosing a nearby $b_t$ for each $a_t$ and taking suprema gives, for every $\pi$,
\begin{equation}
\sup_{P\in\prod_t\mathcal A_t}\Reg_P(\pi)
\le\sup_{Q\in\prod_t\mathcal B_t}\Reg_Q(\pi)+2\sum_tM_td_t.
\label{eq:generic-set-regret}
\end{equation}
Approximate minimizers suffice by taking their error to zero. Minimize over $\pi$, choose the confidence and population sets, and use containment for the lower inequality.
\end{proof}

For a directional enclosure $Q_{t,N,K}\supseteq\widehat\Pset^{\rm CP}_{\Gamma,N,t}$, let $\rho_t$ be its directed distance to the CP set and $\widetilde\Delta$ its optimized regret. Applying \eqref{eq:generic-set-regret} and the triangle inequality gives
\[
0\le\widetilde\Delta-\widehat\Delta^{\rm CP}\le2\sum_tM_t\rho_t,
\qquad
0\le\widetilde\Delta-\Delta_{\rm rect}^M\le2\sum_tM_t(\mathfrak d_t^{\rm CP}+\rho_t),
\]
using coefficients uniform over the enlarged family. Fixed $K$ need not make $\rho_t$ vanish as $N$ increases; this approximation does not inherit the hazard-family rate.

\subsection{Verified projection and numerical evaluation}\label{sec:projected-vertices}\label{sec:numerics}\label{sec:scalable-app}\label{sec:scalable}

The numerical evaluation considered here is restricted to $B\le2$. Confidence levels and $\Gamma$ are represented exactly as rational numbers. Guarantees use costs stored in binary64 (double precision).

\begin{proposition}[One-sided numerical certificate]\label{prop:numerical-certificate-app}\label{lem:dual-directional-projection-app}
Let $Z_t$ be an outward encoding of the latent CP polytope, with demand projection $T Z_t$. Verified dual support bounds define an outer polytope $O_t\supseteq TZ_t$. Suppose its vertices $v_{t,k}$ have stored simplex representatives $\widetilde v_{t,k}$ with $\|v_{t,k}-\widetilde v_{t,k}\|_1\le\eta_t$. For a selected policy, let $U_e^\pi$ and $L_e^*$ be interval-DP upper and lower bounds on its cost and the same-law oracle cost at each stored product $e$. With uniform coefficients over the outer and stored families,
\begin{equation}
\sup_{P\in\prod_tTZ_t}\Reg_P(\pi)
\le U_{\rm pol}+2\sum_tM_t\eta_t,
\qquad U_{\rm pol}=\max_e(U_e^\pi-L_e^*).
\label{eq:numerical-certificate-app}
\end{equation}
On $\mathcal E_N$ this also bounds true-law regret, irrespective of policy optimization accuracy.
\end{proposition}
\begin{proof}
For a count $k$ and tail error $\alpha/2$, nonboundary CP endpoints are certified by exact binomial-tail comparisons:
\[
\Prb_{\widetilde h_L}(X\ge k)\le\alpha/2,\qquad
\Prb_{\widetilde h_U}(X\le k)\le\alpha/2.
\]
Tail monotonicity gives outward enclosure; boundary endpoints are zero or one, and cases beyond the exact-tail verification range return $[0,1]$. A sensitivity row $a^\top x\le0$, $x\ge0$, remains valid after componentwise downward rounding of $a$. Thus $Z_t$ contains the exact CP set.

For each chosen direction $a$, rationally verified dual feasibility and an upward-rounded support value $\overline h(a)$ imply $a^\top Tx\le\overline h(a)$ by weak duality. Exact intersection of these halfspaces with the probability simplex gives $O_t$. Vertices are stored on a dyadic simplex grid with exact unit sum; their maximum error is computed rationally. Applying the same convex weights to stored vertices shows that every point of $O_t$ is within $\eta_t$ of their convex hull, which need not contain $O_t$. Equation~\eqref{eq:generic-set-regret} supplies the storage allowance. Proposition~\ref{prop:vertex-app} reduces the stored-hull supremum to endpoint products, where directed-rounding DP gives $\Reg_e(\pi)\le U_e^\pi-L_e^*$. The allowance uses rationally computed cost spans and valid clearance floors; its final sum is rounded upward. Failed projection or numerical checks prevent issuance of a bound.
\end{proof}

\section{Sample Complexity with Explicit Hidden Selection}\label{sec:gamma-rates}

We prove the bounds for excess over deterministic population debt in \eqref{main-eq:gamma-excess-target}. A learning rule may randomize which policy in $\PiM$ it returns; the adversarial law evaluates that returned policy.

\subsection{Hazard sensitivity and the upper bound}\label{sec:gamma-transform}\label{sec:gamma-upper-proof}

Use the hazard construction of Section~\ref{main-sec:gamma-sample}, with $\beta=\log(4H(3B+2)/\delta)$. Here $\kl(a,b)=a\log(a/b)+(1-a)\log((1-a)/(1-b))$, with $0\log0=0$ and infinite divergence when $b$ assigns zero mass to an event of positive mass under $a$. Denote the cell confidence region by $\mathcal C_{N,t}$, the revealed-hazard interval by $I_{t,d}$, and its transformed interval by $J_{t,d}=[f_\Gamma(h_{L,t,d}),g_\Gamma(h_{U,t,d})]$. Applying \eqref{main-eq:hazard-recursion} to these parameters and taking the convex hull gives $\widehat\Pset^{\mathrm{haz}}_{\Gamma,N,t}$. The untrimmed family is compact and nonempty: the cell region contains the empirical vector and the hazard intervals are nonempty. A known valid clearance restriction may be imposed. If that intersection is empty, return an arbitrary policy and certificate $+\infty$; this cannot occur on coverage.

\begin{lemma}[Hazard confidence-set containment and distance]\label{lem:gamma-set-distance}
For $B,\Gamma\ge1$ and iid historical trajectories, suppose every positive revealed risk is at least $q>0$ and $Nq\ge8\log(2HB/\delta)$. Then, with probability at least $1-\delta$, simultaneously for all $t$,
\begin{equation}
\begin{gathered}
\Pset_{\Gamma,t}\subseteq\widehat\Pset^{\mathrm{haz}}_{\Gamma,N,t},\qquad
\sup_{p'\in\widehat\Pset^{\mathrm{haz}}_{\Gamma,N,t}}\dist_1(p',\Pset_{\Gamma,t})\le d_N,\\
d_N=2\sqrt2(B+1)\sqrt{\frac\beta N}
+8\sqrt2 B\sqrt{\frac{\Gamma\beta}{Nq}}.
\end{gathered}
\label{eq:gamma-set-distance}
\end{equation}
A valid clearance intersection preserves these conclusions.
\end{lemma}

\begin{proof}
\emph{Endpoint sensitivity.}
Put $x(u)=(\sqrt u,\sqrt{1-u})$ and $d_\circ(u,v)=\|x(u)-x(v)\|_2$. For $T=f_\Gamma$ or $g_\Gamma$, $x(T(u))=Ax(u)/\|Ax(u)\|_2$, with $A=\operatorname{diag}(1,\sqrt\Gamma)$ or $\operatorname{diag}(\sqrt\Gamma,1)$. Since $\|Ax(u)\|_2\ge1$, the inequality $\|a/\|a\|-b/\|b\|\|\le2\|a-b\|/\|a\|$ bounds the transformed root distance by $2\sqrt\Gamma\,d_\circ(u,v)$. Cauchy--Schwarz bounds total variation by root distance, and Jensen gives $\kl(u,v)\ge-2\log\langle x(u),x(v)\rangle\ge d_\circ(u,v)^2$. Thus
\begin{equation}
|T(u)-T(v)|\le2\sqrt\Gamma\,d_\circ(u,v),\qquad
|T(u)-T(h)|\le4\sqrt{\Gamma\beta/R}\quad(u,h\in I,\ R>0).
\label{eq:gamma-interval-transform}
\end{equation}
The second inequality uses the triangle inequality through $\widehat h$ and $R\kl(\widehat h,u),R\kl(\widehat h,h)\le\beta$. The formulas extend to zero probabilities by continuity.

\emph{Demand coupling.}
The recursion is a mixture of revealed point masses with weights $o_d$ and censored stopping chains with weights $s_c$. A chain starts at $c$, stops at threshold $d$ with probability $\nu_d$, and terminates at $B$. Changing mixture weights has $\ell_1$ effect at most $\|o'-o\|_1+\|s'-s\|_1$. With weights fixed, couple the chains using the same uniforms: their first separation probability is at most $\sum_d|\nu'_d-\nu_d|$. Terminal $\ell_1$ distance is at most twice that probability. Averaging over censored weights, whose sum is at most one, gives
\begin{equation}
\|p'-p\|_1\le\|o'-o\|_1+\|s'-s\|_1+2\sum_{d<B}|\nu'_d-\nu_d|.
\label{eq:gamma-recursion-coupling}
\end{equation}

\emph{Coverage and projection.}
Each cell interval has failure probability at most $2e^{-\beta}$ by Hoeffding; each nonempty hazard interval has the same bound conditional on its risk count by binomial Chernoff tails. A union bound over $H(3B+2)$ intervals gives coverage with probability at least $1-\delta/2$. For positive population risks, the counts satisfy $R_{t,d}\ge Nq/2$ except on an event of probability at most $HB e^{-Nq/8}\le\delta/2$.

On both events, monotonicity of $f_\Gamma,g_\Gamma$ and Proposition~\ref{main-prop:endpoint-hazard} give containment. For a generated outer point, replace its cells by the true cells, changing their total $\ell_1$ mass by at most $2(B+1)\sqrt{2\beta/N}$. Clip each hidden hazard to its true interval; \eqref{eq:gamma-interval-transform} bounds its change by $4\sqrt{2\Gamma\beta/(Nq)}$. A zero population risk has zero count almost surely and both hidden-hazard intervals equal $[0,1]$, requiring no change. The resulting vector is exact-feasible, and \eqref{eq:gamma-recursion-coupling} gives $d_N$. For convex combinations, use the same weights on their exact-feasible projections; convexity preserves the bound. A valid clearance intersection removes no exact point and cannot increase the directed distance.
\end{proof}

\begin{proof}[Proof of Theorem~\ref{main-thm:gamma-upper}]
On the preceding event, containment and Theorem~\ref{main-thm:contraction} give
\[
0\le\mathcal E_{\Gamma,Q}(\widehat\pi^{\mathrm{haz}}_{\Gamma,N})
\le\widehat\Delta^{\mathrm{haz}}_{\Gamma,N}-\Delta_{\mathrm{rect}}^M(\Gamma)
\le2\sum_tM_td_N\le AHd_N.
\]
Here $2M_t\le S+G/\eta=A$ by the value-span recursion. Since $q\le1$, \eqref{eq:gamma-set-distance} yields \eqref{main-eq:gamma-upper}. If only the exact sets have clearance $\eta_0$, their distance bound gives outer clearance $\eta_0-d_N$; $N\ge1152B^2\Gamma\beta/(q\eta_0^2)$ and the risk-count condition ensure $d_N\le\eta_0/2$. Without uniform clearance, $2M_t\le S+G(H-t)$ gives the stated $H^2$ upper order.
\end{proof}

\subsection{A binary experiment and the lower bound}\label{sec:gamma-lower-proof}

Fix $\Gamma\ge2$, $0<q\le1/2$, and $N\ge1$. For $B=1$, draw $C\in\{0,1\}$ with $\Prb(C=1)=q$ and set $\Prb(D=0\mid C=1)=h$. The full recorded triple has law
\begin{equation}
Q_h(0,0,0)=1-q,\qquad Q_h(0,1,1)=qh,\qquad Q_h(1,0,1)=q(1-h).
\label{eq:gamma-hard-log}
\end{equation}
Its threshold-zero revealed risk is $q$. Given $C=0$, any zero-demand probability in $[f_\Gamma(h),g_\Gamma(h)]$ gives a compatible completion without changing the log. Hence the exact demand-zero interval is
\begin{equation}
[L(h),U(h)]=[qh+(1-q)f_\Gamma(h),\ qh+(1-q)g_\Gamma(h)].
\label{eq:gamma-hard-interval}
\end{equation}
Let $h_*=1/(\Gamma+1)$ and $K_*=1-[L(h_*)+U(h_*)]/2$. Start empty, with unit shortage cost, setup cost $K_*$, and zero holding and per-unit ordering costs. This known cost is common to both alternatives below at fixed $\Gamma,q$, lies in $(0,1)$, and bounds realized stage costs by one.

\begin{lemma}[A hard pair with positive debt]\label{lem:gamma-hard-gap}\label{lem:gamma-hard-kl}
Set
\begin{equation}
a=\frac1{20}\min\left\{\frac1{\Gamma+1},\frac1{\sqrt{Nq(\Gamma+1)}}\right\},
\qquad h_\pm=h_*\pm a,\qquad
\omega_N=\min\left\{1,\sqrt{\frac\Gamma{Nq}}\right\}.
\label{eq:gamma-hard-separation}
\end{equation}
Both experiments have positive debt. Their optimal common actions are no order at $h_+$ and order one at $h_-$. A wrong action has excess worst regret at least $\Gamma a/10$, and every compatible law has $\Prb(D=1)\ge9/20$. In either direction the $N$-record KL divergence is at most $1/50$, the corresponding total variation is at most $1/10$, and $\Gamma a\ge\omega_N/30$.
\end{lemma}

\begin{proof}
The action costs are $1-p_0$ and $K_*$, giving worst regrets
\[
R_0(h)=[1-L(h)-K_*]^+,\qquad R_1(h)=[K_*-1+U(h)]^+.
\]
At $h_*$ these are equal, and $f_\Gamma(h_*)=1/(\Gamma^2+1)$, $g_\Gamma(h_*)=1/2$ imply $U(h_*)-L(h_*)\ge3/20$. Since $L',U'\le\Gamma$ and $\Gamma a<1/20$, the midpoint $1-K_*$ remains strictly inside both perturbed intervals. Both regrets are therefore positive. Also $U(h_\pm)\le1/2+\Gamma a\le11/20$, proving clearance.

For $|h-h_*|\le a$, $h\le21/[20(\Gamma+1)]$ gives $g'_\Gamma(h)\ge\Gamma/(41/20)^2\ge\Gamma/5$. Hence $(L+U)'\ge\Gamma/10$, and
$R_1(h)-R_0(h)=L(h)+U(h)-L(h_*)-U(h_*)$
has opposite signs and magnitude at least $\Gamma a/10$ at $h_\pm$.

For either $r=h_\pm$, $r\ge19/[20(\Gamma+1)]$ and $1-r\ge13/20$, so $r(1-r)\ge1/[2(\Gamma+1)]$. Thus $\kl(p,r)\le(p-r)^2/[r(1-r)]$ gives, in either direction,
\[
\operatorname{KL}(Q_{h_+}^{\otimes N}\|Q_{h_-}^{\otimes N})
=Nq\kl(h_+,h_-)\le8Nq(\Gamma+1)a^2\le1/50.
\]
Pinsker gives total variation at most $1/10$. Finally, $\Gamma/(\Gamma+1)\ge2/3$ and $\Gamma/\sqrt{\Gamma+1}\ge(2/3)\sqrt\Gamma$ give $\Gamma a\ge\omega_N/30$.
\end{proof}

\begin{proof}[Proof of Theorem~\ref{main-thm:gamma-lower}]
Put $m=\lfloor H/2\rfloor$. For $v\in\{-1,+1\}^m$, decision period $2j-1$ has log $Q_{h_*+v_ja}$. Its following reset period has demand one, setup cost $1/2$, and all other costs zero. If $H$ is odd, append a cost-free period with demand one. Reset and final periods have records $C=1$, identifying their demand exactly with revealed risk one. Independent periodwise completions and iid trajectories satisfy A1--A5; all compatible laws have clearance at least $9/20$.

Every reset empties inventory regardless of the policy. Ordering there adds cost without affecting future inventory, so no reset order can improve the learner or oracle. Each decision starts empty. Without reset orders, both policy cost and informed minimum split across decision periods. With $r_{j,a}$ the worst regret for action $a$, rectangularity allows every marginal to maximize its own same-law regret independently. Therefore
\begin{equation}
\Delta_{\mathrm{rect}}^M(\Gamma)=\sum_{j=1}^m\min_a r_{j,a}>0,\qquad
\mathcal E_{\Gamma,Q_v}(\pi)\ge\frac{\Gamma a}{10}Z_v(\pi),
\label{eq:gamma-lower-reset}
\end{equation}
where $Z_v$ counts wrong robust actions. Equality in the separable regret calculation is attained by policies making no reset orders.

Under the uniform prior on $v$, pair alternatives differing only in one coordinate. All other samples have identical laws, so the full-sample total variation remains at most $1/10$ by Lemma~\ref{lem:gamma-hard-kl}. Adjoining an independent random seed preserves this bound. Any learning rule therefore has average testing error at least $(1-1/10)/2=9/20$ per coordinate, giving some $v$ with $\E_vZ_v\ge9m/20$. As $m\ge H/3$,
\[
\E_v\mathcal E_{\Gamma,Q_v}(\widehat\pi)
\ge\frac{\Gamma a}{10}\frac{9m}{20}\ge\frac H{2000}\omega_N.
\]
For this same $v$, $\E_vZ_v\le9m/40+m\Prb_v(Z_v\ge9m/40)$ implies $\Prb_v(Z_v\ge9m/40)\ge9/40$. On that event, \eqref{eq:gamma-lower-reset} gives excess at least $H\omega_N/4000$, proving the weaker constants $1/5000$ and $1/5$ in \eqref{main-eq:gamma-lower-prob}. Expectations and probabilities include the learning rule's random selection of a deterministic deployed policy; the benchmark remains its own experiment's population debt.
\end{proof}

\begin{proof}[Proof of Corollary~\ref{main-cor:gamma-sample}]
For $B,\Gamma\ge1$,
$2\sqrt2(B+1)+8\sqrt2B\sqrt\Gamma\le12\sqrt2B\sqrt\Gamma$.
Squaring Theorem~\ref{main-thm:gamma-upper} and including its risk-count condition gives \eqref{main-eq:gamma-samples-upper}. For $\delta<1/5$ and $\varepsilon<H/5000$, Theorem~\ref{main-thm:gamma-lower} precludes a uniform guarantee whenever $H\omega_N/5000>\varepsilon$, yielding \eqref{main-eq:gamma-samples-lower}. The binary lower family has uniformly bounded costs and obeys the known clearance restriction $p_1\ge1/4$ throughout its exact sets. Imposing that restriction on the hazard outer family fixes its clearance constant independently of $\Gamma,H$. Thus the stated exponents match up to logarithms.
\end{proof}

\section{Experimental Designs and Decision Diagnostics}\label{sec:experiments}

\subsection{Controlled studies}\label{sec:controlled}\label{sec:population-risk}
All controlled studies start empty with $\delta=0.05$. The binary design uses $H=2$, $\Prb(D_t=0)=0.45$, revelation probabilities $(0.4375,0.70)$ for demands $(0,1)$, costs $(K,h,b,c)=(0.10,0.20,0.25,0)$, true sensitivity 3, candidates $1.5,3,5$, and 1,000 datasets per $N=100,250,500,1{,}000$. Per period, Lemma~\ref{lem:binary-coefficient-outer-app} bounds four cells and the revealed tail with five CP intervals, each at $0.05/10$. Regret coverage is 8,000/8,000 for valid candidates (cellwise Wilson intervals $[99.6,100.0]\%$); at $N=1{,}000$, misspecified candidate 1.5 covers regret in 77.0\% ($[74.3,79.5]\%$) and demand in 25.4\%.

The ternary study uses $H=3$, $p=(0.25,0.32,0.43)$, $o=(0.15,0.12,0.08)$, $s=(0.20,0.20,0.25)$, and true sensitivity 5.5. Costs are $K=(0.45,0.50,0.40)$, $c=(0,0,0)$, $h=(0.20,0.25,0.20)$, and $b=(1.20,1.00,1.10)$. For each $N$, 20 independent panels sample $N$ observations per period, shared across candidates $3,5.5,8$. Thirty CP intervals each receive $1/600$. Eight binary64 directions $(\cos(j\pi/4),\sin(j\pi/4))$, $j=0,\ldots,7$, in $(p_0,p_1)$ give eight stored vertices per period and 512 products. Population risks in Figure~\ref{main-fig:pac-computation} use all 216 policies and 64 population endpoint products; directional comparisons are fixed-sample, not replication means.

\paragraph{Hazard--CP comparison.}\label{sec:hazard-cp}
The comparison uses $B=1,H=2$, $\Gamma\in\{2,5,10\}$, $q\in\{0.1,0.3\}$, both signs, and $N\in\{250,1{,}000,4{,}000\}$, with 200 paired datasets per cell. Define
\[
h_*=\frac1{\Gamma+1},\qquad
 a=\frac1{20}\min\left\{\frac1{\Gamma+1},\frac1{\sqrt{1{,}000q(\Gamma+1)}}\right\},\qquad h=h_*\pm a.
\]
The first-period log follows \eqref{eq:gamma-hard-log}, with true hidden hazard $g_\Gamma(h)$ and intervals $[L(h),U(h)]$ from \eqref{eq:gamma-hard-interval}. Set $K_1=1-[L(h_*)+U(h_*)]/2$, $b_1=1$, $K_2=1/2$, and other costs zero. Period two has $D=C=1$. Law and costs remain fixed across $N$; methods share $N$ iid two-period records.

Hazard uses four-cell Hoeffding intervals on the simplex, conditional KL intervals with $\beta=\log800$, and the odds transforms. Failure budgets are $0.025$ for intervals and $0.025$ for the rate proof's risk-count event; containment uses only the first. CP divides $0.05$ across ten two-sided intervals. These budgets bound error probabilities. Both enumerate four policies and endpoint products with outward-rounded endpoints and exact rational evaluation of represented costs. No clearance restriction is added; six $N=250,q=0.1$ cells fail the rate theorem's risk-count gate. Population-set coverage is 7,200/7,200 for hazard and 7,195/7,200 for CP; CP true-law coverage is 7,197/7,200, with matching certificate counts. All five CP misses involve observable-cell noncoverage. Counts pool heterogeneous instances; paired $t$ and Wilson intervals are pointwise by fixed cell.

\subsection{Retail construction and matched comparisons}\label{sec:retail}\label{sec:mechanisms}\label{sec:method-comparison}\label{sec:pooled-dependence}
The strict window $[6:22)$ requires initial stock, at most one persistent stockout transition, and no positive out-of-stock (OOS) hour sales. Eligibility requires 30 naturally uncensored days through November 24, 2024 and a positive training 90th-percentile scale $s_j$. Fixed-seed selection chooses 2,000 of 17,681 eligible series; the proxy is $\operatorname{clip}(\operatorname{rint}(2y/s_j),0,2)$, with ties to even and frozen scales. No path crosses the cutoff.

With $H=B=2$, training paths are sampled with replacement. Capacity calibration uses all 484,628 strict-clean training-day records in the selected 2,000 series, including censored days and days outside two-day paths. Capacities equal 3 on naturally uncensored days and the clipped quantity otherwise, with counts $(28{,}678,88{,}914,42{,}339,324{,}697)$ at $0,1,2,3$. Independently per cell, draw $C^0$ accordingly, $M\sim\mathrm{Bernoulli}(0.65)$ and $\xi\sim\mathrm{Bernoulli}(1/2)$. When $M=1$, replace $C^0$ by $\max\{0,3-D-\xi\}$ (adverse) or $\min\{3,D+\xi\}$ (aligned). Sensitivities are 6.987 and 24.983. Baseline costs are $(K,c,h,b)=(0.50,0,0.20,1.00)$; $N=250,1{,}000$, 30 paired panels, and candidates $1,3,10,30$. Plug-in uses unsmoothed cells; general CP allocates $0.05/20$ per interval. Four methods cross plug-in/CP inference with regret/original-cost optimization over all 36 policies; each candidate uses a shared eight-direction CP assessment set.

The grid reuses 30 adverse $N=1{,}000$ panels for $K\in\{0.1,0.5,1\}$, $h\in\{0,0.2,0.6\}$, $b\in\{0.5,1,2\}$, $c=0$, four methods and candidates $1,10$. All 27 cells use common $\Gamma=10$ population and log-specific CP assessment sets (Table~\ref{tab:cost-grid-all}). For $\Gamma=1$ policies these common-$\Gamma=10$ sets differ from the selection sets. Pointwise paired 95\% $t$ intervals condition on fixed calibration and pools; reusing panels adds no independent samples.

\begin{table}[t]
\centering
\begin{threeparttable}
\caption{Complete cost grid: CP-regret minus plug-in-regret at $\Gamma=10$}
\label{tab:cost-grid-all}
\small
\begin{tabular*}{\textwidth}{@{\extracolsep{\fill}}rrrrrrr@{}}
\toprule
$K$ & $h$ & $b$ & $\Delta\Reg_{P^0}$ & $\Delta(R_{\rm pop}-\Delta_{\rm rect}^M)$ & $\Delta U$ & Changed \\
\midrule
0.1 & 0 & 0.5 & -0.0009 & -0.0009 & -0.0012 & 3.3\% \\
0.1 & 0 & 1 & 0.0000 & 0.0000 & 0.0000 & 0.0\% \\
0.1 & 0 & 2 & 0.0000 & 0.0000 & 0.0000 & 0.0\% \\
0.5 & 0 & 0.5 & 0.0000 & 0.0000 & 0.0000 & 0.0\% \\
0.5 & 0 & 1 & 0.0000 & 0.0000 & 0.0000 & 0.0\% \\
0.5 & 0 & 2 & -0.0314 & +0.0302 & -0.0522 & 70.0\% \\
1 & 0 & 0.5 & 0.0000 & 0.0000 & 0.0000 & 0.0\% \\
1 & 0 & 1 & 0.0000 & 0.0000 & 0.0000 & 0.0\% \\
1 & 0 & 2 & 0.0000 & 0.0000 & 0.0000 & 0.0\% \\
\addlinespace[2pt]
0.1 & 0.2 & 0.5 & -0.0034 & -0.0024 & -0.0054 & 30.0\% \\
0.1 & 0.2 & 1 & +0.0025 & -0.0043 & -0.0123 & 20.0\% \\
0.1 & 0.2 & 2 & 0.0000 & 0.0000 & 0.0000 & 0.0\% \\
0.5 & 0.2 & 0.5 & 0.0000 & 0.0000 & 0.0000 & 0.0\% \\
0.5 & 0.2 & 1 & 0.0000 & 0.0000 & 0.0000 & 0.0\% \\
0.5 & 0.2 & 2 & 0.0000 & 0.0000 & 0.0000 & 0.0\% \\
1 & 0.2 & 0.5 & 0.0000 & 0.0000 & 0.0000 & 0.0\% \\
1 & 0.2 & 1 & 0.0000 & 0.0000 & 0.0000 & 0.0\% \\
1 & 0.2 & 2 & 0.0000 & 0.0000 & 0.0000 & 0.0\% \\
\addlinespace[2pt]
0.1 & 0.6 & 0.5 & 0.0000 & 0.0000 & 0.0000 & 0.0\% \\
0.1 & 0.6 & 1 & +0.0344 & +0.0712 & -0.0100 & 100.0\% \\
0.1 & 0.6 & 2 & -0.0171 & +0.0778 & -0.0425 & 73.3\% \\
0.5 & 0.6 & 0.5 & 0.0000 & 0.0000 & 0.0000 & 0.0\% \\
0.5 & 0.6 & 1 & -0.0176 & -0.0044 & -0.0120 & 16.7\% \\
0.5 & 0.6 & 2 & 0.0000 & 0.0000 & 0.0000 & 0.0\% \\
1 & 0.6 & 0.5 & 0.0000 & 0.0000 & 0.0000 & 0.0\% \\
1 & 0.6 & 1 & +0.0523 & -0.0089 & -0.0157 & 33.3\% \\
1 & 0.6 & 2 & -0.0121 & -0.0031 & -0.0192 & 30.0\% \\
\bottomrule
\end{tabular*}
\begin{tablenotes}\footnotesize
\item \textit{Notes.} Means over 30 common logs at each cost, with $c=0$. Here $\Delta$ before a metric denotes the paired difference. Changed is the fraction with different actions on at least one of the nine demand paths from empty inventory. All cells, including ties and adverse changes, are retained.
\end{tablenotes}
\end{threeparttable}
\end{table}

For high-service costs $(K,c,h,b)=(0.35,0,0.10,2.00)$, 20 adverse $N=1{,}000$ pairs select plug-in regret policies at candidates 1 and 10. With path shares $w_j$, marginals $P_{tj}$ and joint laws $Q_j$, targets are
\[
P_{\rm pool}=\Bigl(\sum_jw_jP_{1j}\Bigr)\otimes\Bigl(\sum_jw_jP_{2j}\Bigr),\quad
P_{\rm mix}=\sum_jw_j(P_{1j}\otimes P_{2j}),\quad
P_{\rm joint}=\sum_jw_jQ_j.
\]
Holding policies fixed, candidate-10 minus candidate-1 cost changes are respectively $-0.0433$ $[-0.0481,-0.0385]$, $0.0031$ $[0.0027,0.0034]$, and $0.0521$ $[0.0464,0.0579]$ (paired 95\% intervals). The 823 series without held-out paths have zero weight. These reversals concern distinct targets, without causal attribution or population-resampling uncertainty.

\subsection{A controlled revelation study}\label{sec:pilot}\label{sec:pilot-hazards}
Here $H=2,B=1$ and both periods share $Q$ with $(o_0,o_1,s_0,s_1)=(0.08,0.32,0.60,0)$, revealed hazard 0.20, and hidden hazard $3/7$ or $3/5$ (sensitivities 3 or 6), under prior $\Gamma\le10$. Costs are
\[
\begin{aligned}
K&=(0.77990923,0.16179104),&c&=(0.39487302,0.10459798),\\
h&=(0.73867419,0.26848217),&b&=(1.72755703,2.95004377).
\end{aligned}
\]
Each of 500 replications per scenario reuses one multinomial vector of 1,000 iid marginal $Q$-observations across periods, without paired trajectories. Future demand is a product; confidence marginals vary separately. An independent stream of 1,600 Bernoulli observations from the baseline-unrevealed group supplies prefixes $n=0,25,50,100,200,400,800,1{,}600$, shared by both methods. Randomized deeper inventory is assumed to reveal unchanged demand. CP/direct intersects the prior set with
\begin{equation}
\underline\nu^{\rm pil}_{t,d}(u_{t,d}+v_{t,d})\le u_{t,d}
\le\overline\nu^{\rm pil}_{t,d}(u_{t,d}+v_{t,d}).
\label{eq:pilot-hazard-rows}
\end{equation}
CP/$\Gamma$ instead takes the largest bidirectional odds ratio over historical and pilot hazard intervals, capped at 10. Six historical intervals receive $0.025/6$ each and the pilot interval 0.025. Shared true parameters give both-period coverage with failure probability 0.05 at each fixed $n$, without optional stopping.

For $a=o_0,c=s_0$, rationally minimize $a+Lc$ and maximize $a+Uc$ over the historical cell bounds and $c\in[1-r_U,1-r_L]$, $1-o_{1,U}-s_{1,U}\le a+c\le1-o_{1,L}-s_{1,L}$, where $r$ is revealed risk. Four policies and four endpoint products suffice. Initial policies are true-law oracles; direct zero bounds at $n=1{,}600$ retain positive-width demand intervals. All 18 negative gains have sensitivity 3: 14 at $n=25$, four at $n=50$.

\subsection{Computational comparisons}\label{sec:computation}
Table~\ref{tab:computation} reports three fresh-process runs per instance for column-and-constraint generation (CCG), all-endpoint optimization, and independent policy-by-policy dynamic programming (DP) (Section~\ref{main-sec:scalable}). Independent enumeration checks 480 CP instances over 36 policies and 49--64 endpoint products; the maximum optimal-value discrepancy is $6.7\times10^{-16}$.

Finite-set agreement checks optimization. Certifying $U$ also requires one-sided endpoint and projection checks, vertex-representation error control, and interval policy/oracle costs (Proposition~\ref{prop:numerical-certificate-app}).

\begin{table}[htbp]
\centering
\begin{threeparttable}
\caption{Computation on two fixed population instances with $B=2$}
\label{tab:computation}
\small
\begin{tabular*}{\textwidth}{@{\extracolsep{\fill}}rrrrrrr@{}}
\toprule
$H$ & Policies & Products & CCG (ms) & Full set (ms) & Enumerate (ms) & Peak (MiB) \\
\midrule
2 & 36 & 16 & 34.015 & 2.578 & 10.163 & 80.65 \\
3 & 216 & 64 & 147.764 & 18.265 & 491.107 & 82.84 \\
\bottomrule
\end{tabular*}
\begin{tablenotes}\footnotesize
\item \textit{Notes.} Medians of three fresh-process runs at $\Gamma=3$, four endpoints per period; all three optimal values agree. Full set uses vectorized policy enumeration; CCG uses occupancy masters with enumeration checks. Peak memory covers the process. Timings compare alternative optimization methods, excluding end-to-end certification. Platform: Intel Core Ultra 5 225H, Windows 11, Python 3.14.6, SciPy 1.18.0. Thread environment variables were one; actual HiGHS threads were unavailable.
\end{tablenotes}
\end{threeparttable}
\end{table}

\end{document}